%% file: acm_v2_19/acmmanuscript.tex
\documentclass[sigconf,pbalance]{acmart}
\AtBeginDocument{%
  }

\usepackage{microtype}
\usepackage{url}
\usepackage{booktabs}
\usepackage{graphicx}
\usepackage{lineno}
\usepackage{enumitem}
\usepackage{listings} %
\usepackage{svg}

\usepackage{multirow}
\usepackage{bigdelim}
\usepackage{todonotes}
\usepackage{longtable}
\usepackage{tabularray}
\usepackage{wrapfig}
\usepackage[most]{tcolorbox} 
\usepackage{url}
\usepackage{xspace}
\usepackage{svg}
\usepackage[absolute]{textpos} %

\usepackage{fdsymbol}   %

\usepackage[utf8]{inputenc} %
\usepackage[T1]{fontenc}    %
\usepackage{url}            %
\usepackage{booktabs}       %
\usepackage{nicefrac}       %
\usepackage{microtype}      %
\usepackage[table]{xcolor}         %
\usepackage[most]{tcolorbox}
\usepackage{csquotes}
\usepackage{wrapfig}
\usepackage{siunitx}
\usepackage{graphicx}
\usepackage{arydshln}
\usepackage{wrapfig}
\usepackage{enumitem}
\usepackage{soul} %

\usepackage{multirow}
\usepackage{xspace}
\usepackage{adjustbox}
\usepackage{pifont}
\usepackage{caption}
\usepackage{makecell}
\usepackage{subcaption}
\usepackage{bold-extra}
\usepackage{url}
\usepackage{float}
\usepackage{pgf-pie}
\usepackage{balance}
\makeatletter
\newcommand{\balancebodycolumns}{%
  \global\let\elastic@normaloutputdblcol\@outputdblcol
  \balance}
\newcommand{\restorebodycolumns}{%
  \global\let\@outputdblcol\elastic@normaloutputdblcol}
\makeatother

\usepackage[normalem]{ulem}

\newif\ifshowchanges
\showchangesfalse

\usepackage{cleveref}
\crefname{figure}{Figure}{Figures}
\crefname{section}{Section}{Sections}
\crefname{equation}{Equation}{Equations}
\crefname{appendix}{Appendix}{Appendice}
\crefname{table}{Table}{Tables}

\usepackage{pifont}%
\usepackage{tikz}

\definecolor{codekw}{rgb}{0.13,0.13,0.55}
\definecolor{codestr}{rgb}{0.0,0.40,0.0}
\definecolor{codecmt}{rgb}{0.45,0.45,0.45}
\lstdefinestyle{tevpy}{
  language=Python,
  basicstyle=\ttfamily\fontsize{7.5}{10}\selectfont,
  keywordstyle=\color{codekw}\bfseries,
  stringstyle=\color{codestr},
  commentstyle=\color{codecmt},
  numbers=none,
  showstringspaces=false,
  breaklines=false,
  columns=fixed,
  basewidth=0.5em,
  keepspaces=true,
  frame=single,
  framerule=0.3pt,
  xleftmargin=3.3pt,
  xrightmargin=3.3pt,
  rulecolor=\color{gray!50},
  framesep=3pt,
  aboveskip=7pt, belowskip=2pt,
}

\newenvironment{elasticcode}
  {\par\addvspace{6pt}\noindent\begin{minipage}{\columnwidth}}
  {\end{minipage}\par\addvspace{6pt}}

\usepackage{nicematrix}
\newcolumntype{L}[1]{>{\raggedright\let\newline\\\arraybackslash\hspace{0pt}}m{#1}}
\newcolumntype{C}[1]{>{\centering\let\newline\\\arraybackslash\hspace{0pt}}m{#1}}
\newcolumntype{R}[1]{>{\raggedleft\let\newline\\\arraybackslash\hspace{0pt}}m{#1}}
\newcolumntype{P}[1]{>{\centering\let\newline\\\arraybackslash\hspace{0pt}}m{#1}}
\newcommand{\xzhichao}[1]{{\color{red} [ZX]: #1}}
\newcommand{\ours}{\texttt{Tevatron-Elastic}\xspace}

\newcommand{\sans}[1]{\textsf{#1}}

\newcommand{\huggingface}{\raisebox{-1.5pt}{\includegraphics[height=1.05em]{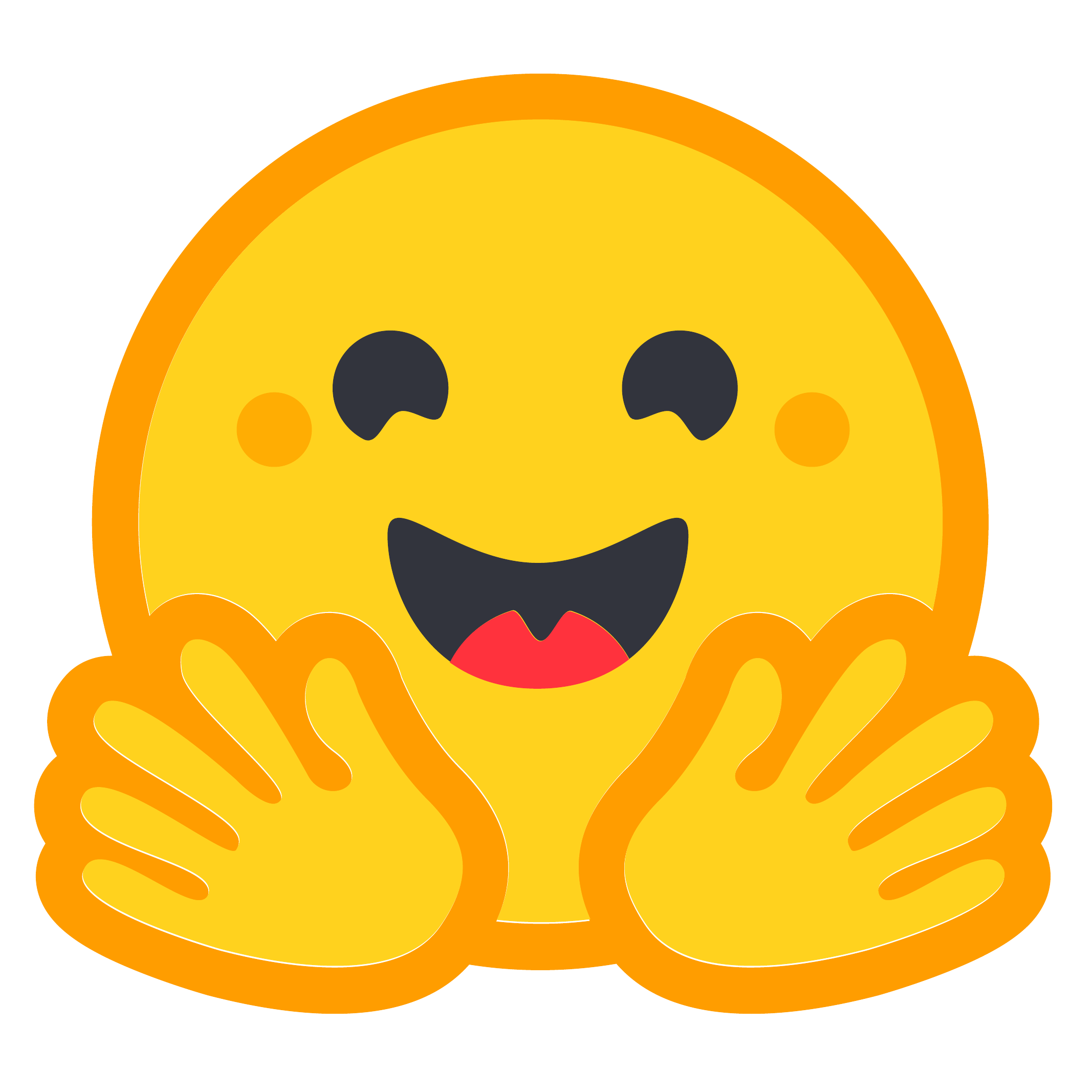}}\xspace}

\newcommand{\github}{\raisebox{-1.5pt}{\includegraphics[height=1.05em]{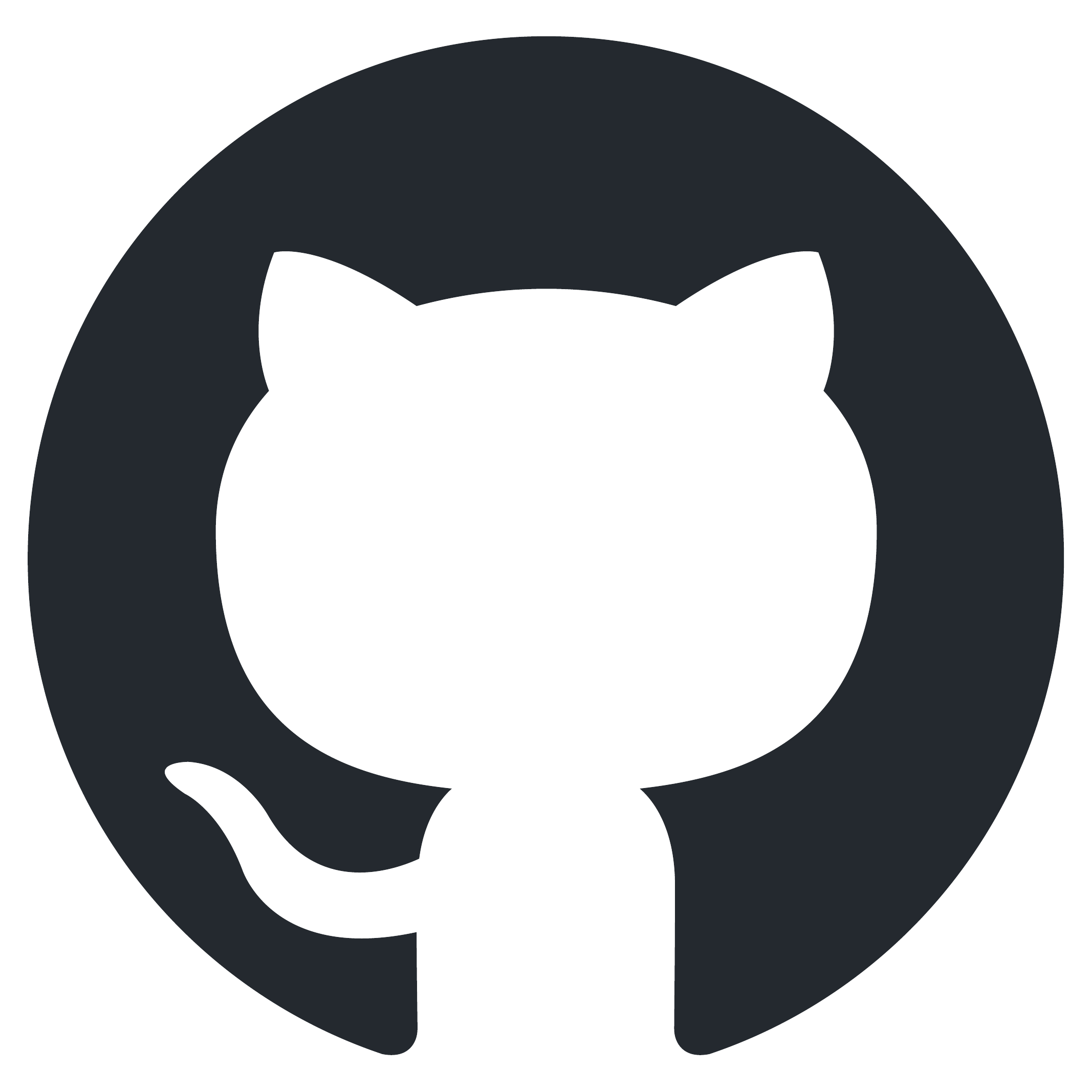}}\xspace}

\copyrightyear{2026}
\acmYear{2026}
\setcopyright{cc}
\setcctype{by}
\acmConference[SIGIR-AP 2026]{Proceedings of the 2026 Annual International ACM SIGIR Conference on Research and Development in Information Retrieval in the Asia Pacific Region}{December 13--15, 2026}{Singapore, Singapore}
\acmBooktitle{Proceedings of the 2026 Annual International ACM SIGIR Conference on Research and Development in Information Retrieval in the Asia Pacific Region (SIGIR-AP 2026), December 13--15, 2026, Singapore, Singapore}
\acmDOI{10.1145/3848000.3850194}
\acmISBN{979-8-4007-3033-7/2026/12}


\begin{document}

\title{Tevatron-Elastic: A Unified Abstraction for Training Elastic Retrievers and Rerankers}

\author{Yu Wang}
\email{yuki.wang@utah.edu}
\affiliation{%
  \institution{University of Utah}
  \city{Salt Lake City}
  \state{UT}
  \country{USA}
}

\author{Shengyao Zhuang}
\email{s.zhuang@uq.edu.au}
\affiliation{%
  \institution{The University of Queensland}
  \city{Brisbane}
  \state{QLD}
  \country{Australia}
}

\author{Xueguang Ma}
\email{x93ma@uwaterloo.ca}
\affiliation{%
  \institution{University of Waterloo}
  \city{Waterloo}
  \state{ON}
  \country{Canada}
}

\author{Zongyu Wu}
\email{zongyuwu@psu.edu}
\affiliation{%
  \institution{The Pennsylvania State University}
  \city{University Park}
  \state{PA}
  \country{USA}
}

\author{Jimmy Lin}
\email{jimmylin@uwaterloo.ca}
\affiliation{%
  \institution{University of Waterloo}
  \city{Waterloo}
  \state{ON}
  \country{Canada}
}

\author{Vivek Srikumar}
\email{svivek@cs.utah.edu}
\affiliation{%
  \institution{University of Utah}
  \city{Salt Lake City}
  \state{UT}
  \country{USA}
}

\author{Zhichao Xu}
\email{zhichao.xu@utah.edu}
\affiliation{%
  \institution{University of Utah}
  \city{Salt Lake City}
  \state{UT}
  \country{USA}
}

\renewcommand{\shortauthors}{Yu Wang et al.}



\begin{abstract}
A single model scale challenges the flexibility of a production retrieval system: some settings need it faster,
others need a smaller index, and the right trade-off changes with the workload. In the context of information retrieval (IR), a transformer-based model can
be made smaller in three ways\,---\,using fewer layers, passing fewer tokens through the upper layers,
or producing a shorter embedding\,---\,and each way saves a different resource. These options have been
studied in isolation, each as its own method with its own code and training setup, which makes
them hard to combine or adapt to a new model. In this work, we present~\ours to bring all three under one
unified abstraction: a single object names any size the model can run at, and a short schedule lists
the sizes to train. Training then produces one checkpoint that serves all of those sizes, and at
deployment the user has the liberty to pick any of them. The same abstraction covers both retrievers and rerankers
and both encoder and decoder models, as it works through interfaces that Hugging Face
transformers already expose; adding a new backbone can be a configuration change rather than a model rewrite.
Prior methods\,---\,Matryoshka embeddings~\citep{kusupati2022matryoshka}, early
exit~\citep{liu-etal-2020-fastbert,xin-etal-2020-deebert}, 2D~Matryoshka (Starbucks)~\citep{zhuang2026starbucks},
and layerwise token compression (LTC)~\citep{Zhuang2026LayerwiseTC}\,---\,become special cases of our unified abstraction.
With \ours, we also propose a new elastic algorithm Matryoshka~LTC (MLTC) to jointly train several token-compression
ratios in one retriever checkpoint.
To validate our framework, we train 20 checkpoints across three backbones and two tasks: the
quality curves are smooth, full-point quality remains close to that of models trained for a single size, and a
controlled study confirms the wall-clock speedups. We release the framework and all checkpoints to support elastic retrieval research.
\end{abstract}


\maketitle

\begin{center}
\small

\begin{minipage}{0.95\columnwidth}
\centering

\github~{\sans{Code:}}~
\href{https://github.com/zhichaoxu-shufe/tevatron-elastic}{\texttt{Tevatron-Elastic}} \\[3pt]

\huggingface~{\sans{Models:}}~
\href{https://huggingface.co/collections/utahnlp/tevatron-elastic}{Tevatron Elastic Collection} \quad

\end{minipage}

\end{center}

\section{Introduction}
\label{sec:intro}

A transformer-based information retrieval (IR) model~\citep{lin2022pretrained,xu2025surveymodelarchitecturesinformation} has three largely separable axes that
control its cost at inference time:
\begin{itemize}[leftmargin=*]
  \item \textbf{depth}\,---\,how many transformer layers are executed before a readout. Fewer layers mean less encoder/scorer compute and lower latency~\citep{devlin-etal-2019-bert,liu-etal-2020-fastbert,xin-etal-2020-deebert,warner2024modernbert}.
  \item \textbf{token}\,---\,how many sequence positions flow through the upper layers. Pooling the sequence partway up the stack reduces the floating-point operations (FLOPs) of those upper layers~\citep{goyal2020powerbert,zhang2025jaspertokencompression600mtechnicalreport}.
  \item \textbf{width}\,---\,the dimensionality of the output embedding. A smaller embedding reduces the size of the stored index and the cost of nearest-neighbor search~\citep{johnson2017billionscalesimilaritysearchgpus,kusupati2022matryoshka}.
\end{itemize}
\noindent A useful mental model to group these axes is by the deployment cost each one reduces: depth and token reduce
\emph{compute}, while width reduces \emph{storage and search}. Depth and token act at compute time
and leave the index unchanged; width acts at storage time and leaves the per-input compute
unchanged.

Prior work on elastic modeling tends to commit to a single axis. Matryoshka representation learning (MRL) varies width~\citep{kusupati2022matryoshka};
layer-dropping and early-exit methods vary depth~\citep{liu-etal-2020-fastbert,xin-etal-2020-deebert,Elhoushi2024layerskip}; \citet{zhang2025jaspertokencompression600mtechnicalreport} and \citet{Zhuang2026LayerwiseTC} compress the retriever and reranker on the token axis; and 2D~Matryoshka retrievers such as 2DMSE and Starbucks vary depth and width
together~\citep{li20242dmatryoshkasentenceembeddings,zhuang2026starbucks}. Each of the aforementioned methods is effective in its own setting, but each comes as a separate implementation with
its own model class definition and training objective, so combining them, or adapting one to a new task or backbone, requires re-implementation.

We posit that these compression axes are complementary rather than competing: because they reduce different costs, they can be applied independently and composed. Existing work, however, provides no single abstraction to unify the definitions of different compressed model classes, as well as the corresponding training and inference.
This brings us to our central research question:

\vspace{3pt}
\begin{center}
\begin{minipage}{0.92\columnwidth}
\itshape Can we name and train these choices in a single framework, while retaining the deployment trade-offs each one offers?
\end{minipage}
\end{center}
\vspace{3pt}

\noindent In this work, we present~\ours to provide a unified abstraction of these axes based on \texttt{Tevatron}~\citep{ma2025tevatron20}. 
We first abstract the action of compressing a model as an \emph{operating point}, e.g., early exit at a certain layer means defining the early exit operating point for that layer.
Based on this definition, \ours introduces two core components. A frozen dataclass, \texttt{Granularity}, names any operating point with fields for depth, width, and token compression; and a \texttt{GranularitySchedule} lists the operating points to train the underlying elastic model. Specifically, the training loop sums a per-granularity loss over the schedule from a shared backbone forward pass and any token-specific upper-layer passes, producing one checkpoint that serves every listed operating point. Instead of editing the model internals, the abstraction routes through
the hidden-states tuple, the layer module list, and the pooling step exposed by Hugging Face transformers. Depth and width need only configuration changes for a new backbone; token compression additionally needs a per-family upper-layer runner.

Our core contribution is the unified abstraction that accommodates different types of compression strategies:
\begin{enumerate}[leftmargin=*]
  \item one abstraction (\cref{sec:framework}) covers retrieval and reranking across three compression axes and five families (BERT, ModernBERT, Qwen3, Llama3, and Mistral) without rewriting model classes;
  \item the same abstraction expresses prior elastic methods: MRL, early exit, 2D~Matryoshka (Starbucks), and LTC as configurations, and recovers their characteristic quality--cost curves under a single shared training recipe (\cref{sec:experiments});
  \item the operating points trained jointly behave as expected\,---\,the quality curves from our empirical studies are smooth and full-point quality remains close to that of dedicated single-point models (\cref{sec:experiments}). Controlled measurements show that inference speedups follow the analytic cost model (\cref{sec:efficiency}).\looseness=-1
  \item a new elastic algorithm \emph{Matryoshka Layerwise Token Compression} (MLTC), which jointly trains several token-compression ratios in one checkpoint, extending prior token compression methods~\citep{Zhuang2026LayerwiseTC,zhang2025jaspertokencompression600mtechnicalreport}. MLTC uses the framework's existing schedule and token readout path; supporting several mid-stack pooling points would otherwise require modifying the model class.
\end{enumerate}

We do not claim that \ours improves on prior methods in absolute quality,
nor that combining all three axes is preferable to using one. In our empirical experiments, we hold the training recipe fixed
(supervised fine-tuning from base checkpoints, without the pre-training some
prior systems use~\citep{gao-callan-2022-cocondenser,zhuang2026starbucks,xu2026laconic}), so that we can study the trained elastic checkpoints in isolation. We release our framework and checkpoints to help researchers reproduce existing elastic model recipes and develop new ones.

\section{Related Work and Background}
\label{sec:background}
We review IR compression and the broader model-compression literature~(\cref{subsec:related-works}), then define the notation used throughout this paper~(\cref{subsec:notations}).

\subsection{Prior methods}
\label{subsec:related-works}

We organize prior work by compression axis, focusing on retrieval or ranking models that are \emph{elastic}\,---\,adjustable in cost after training.

\paragraph{Width.} Matryoshka representation learning trains an embedding whose leading
coordinates remain usable on their own, so one model serves several vector sizes and, in
retrieval, several index sizes~\citep{kusupati2022matryoshka}. The truncation is at the output, so
width changes storage and nearest-neighbor search cost~\citep{johnson2017billionscalesimilaritysearchgpus}
while leaving per-input computation unchanged.

\paragraph{Depth.} Early-exit and layer-dropping methods read out the representations at an intermediate layer to
trade quality for fewer layers of compute, either with a fixed exit or an input-adaptive
one~\citep{liu-etal-2020-fastbert,xin-etal-2020-deebert,Elhoushi2024layerskip}. In IR, the resulting encoder or reranker executes fewer transformer layers.

\paragraph{Token.} Sequence-compression methods shorten the token sequence inside the stack so the
upper layers process fewer positions~\citep{goyal2020powerbert}. Prior works Jasper-Token-Compression-600M~\citep{zhang2025jaspertokencompression600mtechnicalreport} and layerwise token compression (LTC)~\citep{Zhuang2026LayerwiseTC} pool the sequence of tokens at a chosen layer and run the remaining layers on the shortened sequence for retrieval and reranking, respectively.

\paragraph{Composition.} A few methods combine axes. 2D~Matryoshka embeddings vary depth and width
together~\citep{li20242dmatryoshkasentenceembeddings}, and Starbucks trains a retriever elastic on
both, so one checkpoint serves a grid of (layer, dim) points~\citep{zhuang2026starbucks}. These are
the closest prior systems to ours; each, however, fixes one particular pair of axes and one task.

\paragraph{Relation to general model compression.} The three axes above are the IR-specific face of a broader model-compression literature, which we group into two lines. The first produces a single smaller model: structured pruning removes layers, heads, or channels~\citep{xia2024shearedllama,xu-etal-2024-beyond-perplexity}, distillation trains a compact student from a larger teacher~\citep{hinton2015distillingknowledgeneuralnetwork,sanh2019distilbert}, and quantization lowers the numerical precision of weights and activations~\citep{dettmers2022llmint8,frantar2023gptq,lin2024awq}; in IR, distilling cross-encoders into cheaper rankers~\citep{hofstatter2020marginmse,xu2025distillationversuscontrastivelearning} and efficiency-oriented learned sparse retrieval~\citep{formal2022efficientsplade,xu2025cspladelearnedsparseretrieval,xu2026laconic} follow this pattern. The second line trains one network that can be executed at several sizes after training: slimmable and once-for-all networks vary width and depth in convolutional models~\citep{yu2019slimmable,cai2020onceforall}, LayerDrop and DynaBERT do so for transformers~\citep{fan2020layerdrop,hou2020dynabert}, MatFormer and Flextron nest sub-models inside large language models~\citep{devvrit2023matformer,cai2024flextron}. Token reduction sits between the two lines: PoWER-BERT learns a fixed retention configuration and retrains for it~\citep{goyal2020powerbert}, whereas token merging applies at inference time without retraining~\citep{bolya2023tome}; the token axis in our framework instead trains several keep ratios into one checkpoint. \ours belongs to the second, train-once-serve-many line. It is a domain-specific adaptation of these ideas rather than a new compression operator: the depth, token, and width axes are the counterparts of layer, sequence, and width elasticity for the two readouts that IR models use, a pooled embedding and a relevance score. Weight quantization, from the first line, acts on the stored weights and can be applied to any operating point of an elastic checkpoint, so it is complementary to the abstraction we propose; distillation likewise enters the framework as a per-granularity objective rather than as a competing method (\cref{sec:framework}).

\paragraph{Our position.} In existing works, each axis (or fixed pair) is realized as a separate
method with its own model class definition, implementation, and training strategy, and almost always on a
single task. We do not introduce a new pooling operation; instead, we provide one unified abstraction in which
depth, token, and width are named uniformly, the set of operating points to train is written as
data, and the methods above\,---\,MRL, early exit, LTC, and 2D~Matryoshka\,---\,appear as
configurations for both retrieval and reranking. The per-operating-point objective is itself
pluggable and can be a standard contrastive loss, a learning-to-rank
loss~\citep{cao2007learningtorank,xia2008listwiseapproach,oord2018representation}, or
distillation~\citep{hinton2015distillingknowledgeneuralnetwork,xu2025distillationversuscontrastivelearning}
(\cref{sec:framework}). MLTC is one new configuration enabled by this interface: it applies the
same token-pooling path to several ratios in a single retriever checkpoint.

\subsection{Notation}
\label{subsec:notations}

We set up notation shared by the three axes, which \cref{sec:framework} then unifies. A
transformer with $L$ layers maps an input of $T$ tokens to a sequence of hidden states; write
$H^{(\ell)} \in \mathbb{R}^{T \times D}$ for the output of layer $\ell$, with $D$ the hidden size.
A dense encoder pools the top layer to a vector $z = \operatorname{pool}(H^{(L)}) \in \mathbb{R}^{D}$
(CLS, mean, or last token); a cross-encoder maps $H^{(L)}$ to a scalar relevance score. An
operating point specifies the layer, token count, or width for a readout.

\paragraph{Width (MRL)} Width keeps the first $d \le D$ coordinates of the pooled vector,
$z_{1:d}$, and scores with the truncated embedding. Training sums the losses for
all requested widths $\mathcal{D} = \{d_1, \dots, d_k\}$, evaluating the same objective
on each nested prefix~\citep{kusupati2022matryoshka}:
\begin{equation}
\mathcal{L}_{\text{MRL}} = \sum_{d \in \mathcal{D}} \mathcal{L}\big(z_{1:d}\big).
\end{equation}
The forward pass is unchanged; only the output dimension varies, so width affects index size and
search cost, not per-input compute.

\paragraph{Depth (early exit)} Depth reads out at an intermediate layer $\ell \le L$ instead of
the top, using $\operatorname{pool}(H^{(\ell)})$ (or the scalar head applied at layer $\ell$).
Training over a set of exit layers $\mathcal{E} = \{\ell_1, \dots, \ell_k\}$ gives
\begin{equation}
\mathcal{L}_{\text{depth}} = \sum_{\ell \in \mathcal{E}} \mathcal{L}\big(\operatorname{pool}(H^{(\ell)})\big).
\end{equation}
At inference, an exit at layer $\ell$ executes only those $\ell$ layers, giving an approximate relative compute cost of $\ell/L$.

\paragraph{Token (LTC)} Token compression pools the sequence at a chosen layer $p$, reducing its
length from $T$ to $m = \max(1, \operatorname{round}(rT))$ in our implementation for a keep-ratio $r \in (0, 1]$, then runs the remaining
layers $p{+}1, \dots, L$ on the shortened sequence~\citep{Zhuang2026LayerwiseTC}. Writing
$\operatorname{pool}_{\text{seq}}$ for the sequence pooling and $f_{p:L}$ for the upper-layer stack,
\begin{equation}
\tilde H^{(p)} = \operatorname{pool}_{\text{seq}}\!\big(H^{(p)}, r\big), \qquad
z = \operatorname{pool}\big(f_{p:L}(\tilde H^{(p)})\big).
\end{equation}
The layers below $p$ still run on the full sequence, so the compute saving applies only to the upper layers and is largest for long inputs.

Although the three choices intervene at different components of transformer computations, they all read from or transform the
same hidden-state sequence. Naturally, this shared structure suggests a common interface, which we illustrate next in~\cref{sec:framework}.


\section{Framework Design}
\label{sec:framework}

To turn this shared structure into a framework, we first clarify the concept of 
residual stream and the readout. We then describe the requirements for a unified elastic framework.

\paragraph{Preliminary: residual stream and readout.} A backbone processes an input and returns a sequence
of hidden representations as output. This sequence is referred to as the \emph{residual stream} in the literature~\citep{elhage2021mathematicalframework}.
When we set \texttt{output\_hidden\_states=True}, the standard Hugging Face interface returns it as
a tuple $(H^{(0)}, H^{(1)}, \dots, H^{(L)})$, where $H^{(0)}$ is the embedding output and
$H^{(\ell)} \in \mathbb{R}^{T \times D}$ is the output after layer $\ell$. A \emph{readout} is
the method we use to turn this residual stream into the model output at the operating point: choose a layer, optionally shorten its token
sequence, then pool it to a retrieval vector or map it to a relevance score. For a retriever, the
readout can also keep only part of the vector. Depth chooses
\emph{which} layer to read; token compression changes the sequence length $T$; and width truncates
the feature dimension $D$.
\begin{elasticcode}
\begin{lstlisting}[style=tevpy]
out = backbone(**inputs, output_hidden_states=True)
H = out.hidden_states        # stream: H0, ..., HL
h = H[layer]                 # depth readout: [T, D]
z = pool(h, attention_mask)  # cls / mean / eos -> [D]
z = z[:dim]                  # width: first dim
\end{lstlisting}
\end{elasticcode}
\noindent Token compression is the one case that changes the stream \emph{between} layers: it pools
$H^{(p)}$ along $T$ and runs the upper layers on the shorter sequence. The remaining operations
are simple choices in the readout above. This view is the bridge between the three axes.

\paragraph{Desiderata.} Starting from this common readout, a unified framework should meet four
practical requirements:
\begin{enumerate}
\item Name every operating point in one convention, whether it changes depth, token count,
or embedding width.
\item Let users combine operating points without new modeling code, and select the
points to train in a small configuration rather than in a new training loop.
\item Work with the interfaces that transformer models already expose, so adding a backbone requires only configuration changes instead of rewriting the elastic model.
\item Train and serve many operating points efficiently, so users can deploy one checkpoint at different sizes rather than train and maintain a separate model for each size.
\end{enumerate}
\noindent The rest of this section answers these four goals in turn.

\begin{elasticcode}
\paragraph{One description for every operating point.} A frozen dataclass specifies the exit layer, with optional width and token settings:
\begin{lstlisting}[style=tevpy]
@dataclass(frozen=True)
class Granularity:
    layer: int                       # readout / exit layer
    dim: int | None = None           # width: first dim
    keep_ratio: float | None = None  # token: fraction kept
    pool_layer: int | None = None    # token: pool here
\end{lstlisting}
\end{elasticcode}
\noindent A \texttt{Granularity} prints as a compact, collision-free key. Examples include \texttt{L12}, \texttt{L6xD128}, \texttt{L20xR0.8}, and \texttt{L28xR0.6xP20}. The key doubles as the operating point's name in logs and on disk. Fields are independent: setting only \texttt{layer} gives depth, adding \texttt{dim} gives
width, and adding \texttt{keep\_ratio} with \texttt{pool\_layer} gives token compression. The reranker's legacy token path instead uses \texttt{layer} for the pooling point and reads out at the final layer; \texttt{L20xR0.6}, for example, pools at layer 20 and exits at layer 28 on Qwen3-0.6B.

\paragraph{A schedule instead of custom training code.} The set of operating points to train is a list, written as a short
spec string or JSON:
\begin{elasticcode}
\begin{lstlisting}[style=tevpy]
schedule = GranularitySchedule.from_spec("2:32,6:128,12:768")
\end{lstlisting}
\end{elasticcode}
\noindent \cref{tab:spec-grammar} maps spec strings to prior methods. The grammar covers single axes and their compositions, so selecting operating points requires only configuration changes.

\begin{table}[t]
\centering
\small
\caption{Schedule grammar: configurations for individual compression axes, their compositions, and prior methods.}
\label{tab:spec-grammar}
\resizebox{\columnwidth}{!}{
\begin{tabular}{lll}
\toprule
what to train & spec / preset & reproduces \\
\midrule
depth-only (early exit)    & \texttt{"2,4,6,8,10,12"} / \texttt{pure\_depth} & early exit \\
width-only (MRL)           & \texttt{pure\_mrl(12,(32,...,768))}            & MRL \\
depth $\times$ width       & \texttt{starbucks\_bert} / \texttt{diagonal(...)} & Starbucks \\
token-only (reranker)      & \texttt{"20@0.6"}                               & LTC \\
token-only (retriever)     & \texttt{"28@1.0/20,...,28@0.4/20"}             & MLTC (this work) \\
width + token              & \texttt{"28:256@0.6/20"}                       & \,---\, (composition) \\
full three-axis            & \texttt{"24:256@0.6/20"}                       & left open (\cref{sec:experiments}) \\
\bottomrule
\end{tabular}
}
\end{table}

\paragraph{How readout operates on the standard Hugging Face interface.} Because the readout only uses
the hidden-states tuple, the layer module list, and a pooling step\,---\,all standard across Hugging
Face transformers\,---\,we can use one function to realize all three axes for any backbone, parameterized by a
\texttt{Granularity} \texttt{g}:
\begin{elasticcode}
\begin{lstlisting}[style=tevpy]
def _readout(self, hidden_states, attention_mask, g):
    if g.pool_layer is not None:   # token: pool mid-stack
        h_p = hidden_states[g.pool_layer]
        pooled_h, mask = pool_sequence(
            h_p, attention_mask, g.keep_ratio)
        h = self._range_runner(
            self.encoder, pooled_h, mask,
            g.pool_layer, end=g.layer)
    else:                          # depth: read at g.layer
        h, mask = hidden_states[g.layer], attention_mask
    reps = self._pooling(h, mask)  # cls / mean / eos
    if g.dim is not None:          # width: truncate
        reps = reps[:, :g.dim]
    return F.normalize(reps) if self.normalize else reps
\end{lstlisting}
\end{elasticcode}
\noindent The token path alone needs a per-family upper-layer runner to rebuild attention masks and decoder rotary position embeddings for the shortened sequence:
\begin{elasticcode}
\begin{lstlisting}[style=tevpy]
@register_upper_runner("qwen3", "qwen2", "mistral", "llama")
# end=None runs the upper layers to the top.
def _decoder_upper(backbone, hidden, mask, l_star,
                   end=None): ...
\end{lstlisting}
\end{elasticcode}

\paragraph{Matryoshka Layerwise Token Compression (MLTC)}
MLTC extends LTC to jointly train several token-compression ratios in one retriever checkpoint.
Its \texttt{GranularitySchedule} lists the ratios, and the training loss sums their individual losses, as in MRL.

All ratios share the token readout, so MLTC adds no model path. The token runner and schedule support several mid-stack pooling points without modifying Hugging Face model classes. All five families in \cref{tab:backbones} support depth and width; token compression needs a per-family runner and excludes ModernBERT.

\begin{table}[t]
\centering
\small
\caption{Adding a backbone is a launch flag, not new model code. The pooling column is the only
per-backbone choice for the depth and width axes.}
\label{tab:backbones}
\resizebox{\columnwidth}{!}{
\begin{tabular}{llll}
\toprule
backbone & family & pooling & new model code \\
\midrule
BERT-base~\citep{devlin-etal-2019-bert}       & encoder & cls & \,---\, \\
ModernBERT-base~\citep{warner2024modernbert} & encoder & cls & \,---\, \\
Qwen3-0.6B~\citep{yang2025qwen3technicalreport}      & decoder (GQA) & eos & \,---\, \\
Llama3-\{1B,3B,8B\}~\citep{grattafiori2024llama3technicalreport} & decoder (GQA) & eos & \,---\, (full path verified) \\
Mistral-7B-v0.3~\citep{Jiang2023Mistral7B} & decoder (GQA) & eos & \,---\, (full path verified) \\
\bottomrule
\end{tabular}
}
\end{table}

\paragraph{Joint training.} Each tower runs one shared backbone forward pass per batch. We compute
a readout for each operating point; token points also run the upper layers on a pooled sequence.
The training loss sums the per-granularity losses:
\begin{elasticcode}
\begin{lstlisting}[style=tevpy]
# One shared forward pass per tower.
out_q = model(**query, output_hidden_states=True)
out_p = model(**passage, output_hidden_states=True)
loss = sum(
    gran_loss(out_q, out_p, g) for g in schedule.points)
\end{lstlisting}
\end{elasticcode}
\noindent The per-granularity loss \texttt{gran\_loss} is a pluggable function of the readouts, not part of
the training framework: the schedule decides \emph{which} operating points to train, while \texttt{gran\_loss}
decides \emph{how} each is supervised. In our experiments it is the task's standard objective. For
the retriever it is in-batch InfoNCE~\citep{oord2018representation} over the query and passage readouts, which is softmax
cross-entropy over each query's candidate list together with in-batch negatives~\citep{xiong2021ance,qu-etal-2021-rocketqa}\,---\,equivalently, the ListNet objective~\citep{cao2007learningtorank} with a single relevant item:
\begin{elasticcode}
\begin{lstlisting}[style=tevpy]
def gran_loss(out_q, out_p, g):  # InfoNCE / top-1 ListNet
    # q: [B,d]; p: [B*n,d]; scores: [B,B*n].
    q = self._readout(out_q.hidden_states, q_mask, g)
    p = self._readout(out_p.hidden_states, p_mask, g)
    scores = (q @ p.T) / self.temperature
    target = torch.arange(B, device=q.device) * n
    return F.cross_entropy(scores, target)
\end{lstlisting}
\end{elasticcode}
\noindent In every case the schedule and the readout are unchanged; replacing
ListNet with ListMLE or a distillation loss is a matter of plugging in a different \texttt{gran\_loss}.
Because \texttt{gran\_loss} only sees the readouts for an operating point, substituting a different
objective is a local change to this one function. For an example from the learning-to-rank literature, we use ListMLE~\citep{xia2008listwiseapproach}. This objective replaces the ListNet term with the Plackett--Luce likelihood of the target ordering:\looseness=-1
\begin{elasticcode}
\begin{lstlisting}[style=tevpy]
def gran_loss(out_q, out_p, g):           # ListMLE
    q = self._readout(out_q.hidden_states, q_mask, g)
    p = self._readout(out_p.hidden_states, p_mask, g)
    scores = (q @ p.T) / self.temperature
    return listmle(scores, target_order)  # Plackett-Luce
\end{lstlisting}
\end{elasticcode}
\noindent Distillation instead supervises each operating point with a teacher model~\citep{hinton2015distillingknowledgeneuralnetwork,xu2025distillationversuscontrastivelearning}, where
\texttt{score\_with\_teacher} returns the teacher's query--passage scores for the batch~\citep{xu2025distillationversuscontrastivelearning}:
\begin{elasticcode}
\begin{lstlisting}[style=tevpy]
def gran_loss(out_q, out_p, g):  # distillation
    q = self._readout(out_q.hidden_states, q_mask, g)
    p = self._readout(out_p.hidden_states, p_mask, g)
    student = (q @ p.T) / self.temperature
    teacher = score_with_teacher(query, passage)
    return F.kl_div(
        student.log_softmax(-1), teacher.softmax(-1),
        reduction="batchmean")
\end{lstlisting}
\end{elasticcode}
\noindent The reranker follows the same pattern with grouped cross-entropy over each query's candidate list
(the positive at index~0). The training cost of a schedule follows from the same structure. The backbone forward and backward passes are shared by all operating points; a depth or width point adds only a pooling or truncation step and one loss term, whereas a token point also re-runs the layers above its pooling layer on the shortened sequence, so it adds a fraction of a forward pass, the matching backward pass through those layers, and the activations retained for it. Measuring training time and peak memory as the schedule grows remains future work.

\paragraph{Checkpoint serving.} We load the full checkpoint and prune in place to a chosen operating point.
Depth pruning physically removes the upper layers, so the compute saving applies to every input;
width and token are applied at readout time:
\begin{elasticcode}
\begin{lstlisting}[style=tevpy]
model = ElasticDenseModel.load(ckpt)  # or ElasticReranker
model.prune_to(Granularity(16))       # select a point
\end{lstlisting}
\end{elasticcode}
\noindent We store a single checkpoint to serve all trained operating points.\looseness=-1

\paragraph{A cost model for the three axes.} A linear approximation maps a \texttt{Granularity} to its
relative encode/score FLOPs: $\ell/L$ for a depth point at layer $\ell$, and
$(p + (L-p)\,r)/L$ for a token point that pools at layer $p$ with keep-ratio $r$. The token estimate assumes cost linear in sequence length and ignores pooling overhead.
Relative index size is $d/d_{\max}$ for width $d$. \cref{sec:efficiency} checks the measured wall-clock speedup and index size reduction.

\raggedbottom
\section{Experiments}
\label{sec:experiments}
This section presents the experimental setup and results, examining whether the proposed framework
meets the desiderata above.

\subsection{Experimental setup}
\label{sec:setup}

\begin{figure*}[t]
\centering
\captionsetup{skip=6pt}
\includegraphics[width=\textwidth]{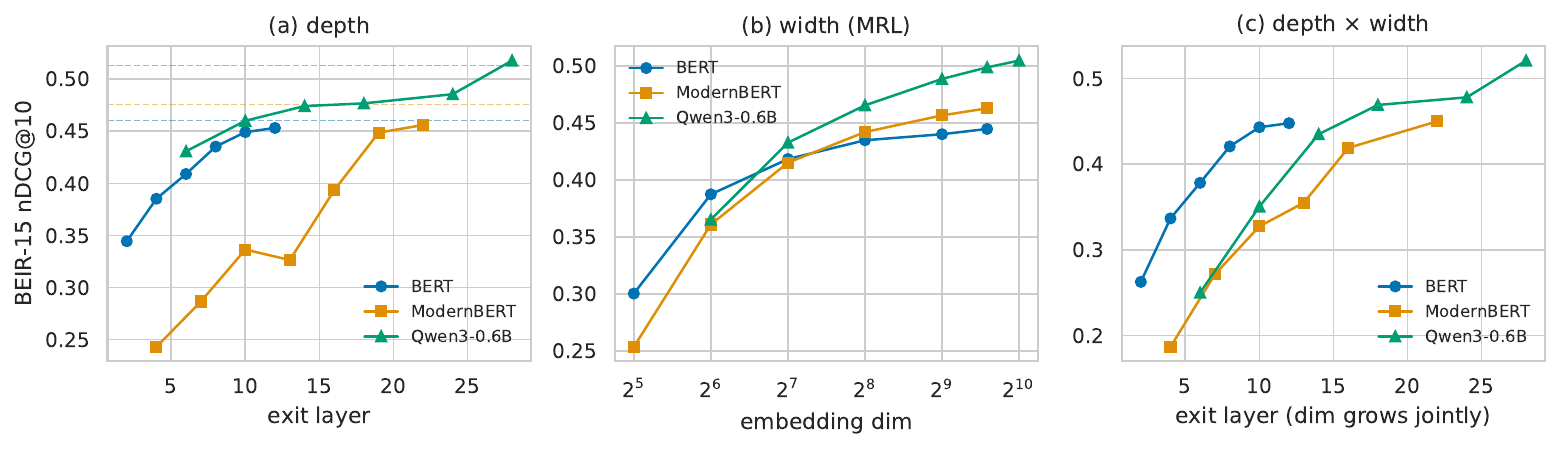}
\caption{Retriever quality (BEIR-15 nDCG@10) along each axis, three backbones. Dashed lines mark
each backbone's plain full point. (a) depth at full width; (b) width (MRL) at full depth, log
dimension axis; (c) depth$\times$width (Starbucks), where layer and dimension grow jointly. Every
curve is smooth except for ModernBERT's step near layer 13 in (a), and the elastic full points sit close to the plain reference.}
\Description{Three panels plot retriever quality against exit layer, embedding dimension, and joint depth-width settings for BERT, ModernBERT, and Qwen3. Dashed lines show the plain full-model references.}
\label{fig:retr-axes}
\end{figure*}

\begin{figure*}[!t]
\captionsetup{skip=6pt}
\begin{minipage}[t]{\columnwidth}
\centering
\includegraphics[width=0.87\linewidth]{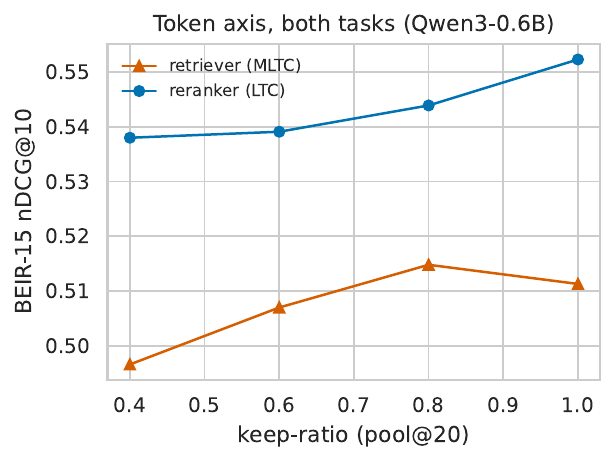}
\caption{Token axis on both tasks (Qwen3-0.6B): the retriever (MLTC) and reranker (LTC) curves
have similar gentle shapes under the same pooling primitive.}
\label{fig:token}
\end{minipage}\hfill
\begin{minipage}[t]{\columnwidth}
\centering
\includegraphics[width=0.87\linewidth]{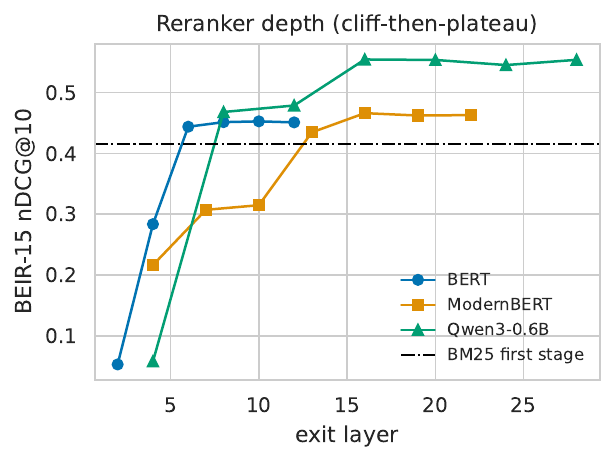}
\caption{Reranker depth, three backbones. A floor at the shallow end then a long plateau, distinct
from the gradual retriever depth curve in \cref{fig:retr-axes}a.}
\label{fig:rerank-depth}
\end{minipage}
\Description{The left figure shows retriever MLTC and reranker LTC quality versus token keep ratio for Qwen3, with pooling at layer twenty. The right figure shows reranker quality versus exit layer for three backbones, with a BM25 reference line; quality rises sharply from shallow exits and then approaches a plateau.}
\end{figure*}

\paragraph{Backbones.} We experiment with three backbone models: BERT-base~\citep{devlin-etal-2019-bert} (110M, encoder,
CLS pooling), ModernBERT-base~\citep{warner2024modernbert} (150M, encoder, CLS pooling, alternating global/local attention), and
Qwen3-0.6B~\citep{yang2025qwen3technicalreport} (decoder, GQA, last-token pooling). Depth and width are evaluated on all three; token experiments use Qwen3 only. Llama3 family models~\citep{grattafiori2024llama3technicalreport} and Mistral-7B~\citep{Jiang2023Mistral7B} are used only to verify the
backbone-agnostic path end to end.

\paragraph{Data and evaluation.} All models are trained on RLHN-680K~\citep{thakur-etal-2025-hard} with \texttt{query:} and
\texttt{passage:} prefixes, at input length 512, despite using different tokenizers. We evaluate on BEIR-15~\citep{thakur2021beir} as a single macro average
(cqadupstack collapsed to the mean of its 12 subforums and counted once; MS~MARCO dev included as
one member), reporting nDCG@10. Retrieval reports absolute nDCG@10; reranking reports
nDCG@10 after reranking BM25 top-100, so the BM25 first stage (0.415) is the reference. The
per-backbone \emph{plain} (non-elastic, single full point) model is the baseline against which we
read each compression curve. 

\begin{table*}[t]
\centering
\caption{Complete per-dataset BEIR-15 results (nDCG@10) for Qwen3-0.6B retrievers and rerankers.}
\label{tab:ds-qwen-all}
\resizebox{\textwidth}{!}{%
\begin{tabular}{lcccccccccccccccc}
\toprule
Model & ARG & CFE & DBP & FEV & FIQ & HOP & NFC & NQ\phantom{0} & QUO & SCD & SCF & TRC & TOU & MSM & CQA & \textbf{AVG} \\
\midrule
\multicolumn{17}{@{}l}{\textit{Qwen3-0.6B retriever, plain.}} \\
28:1024 & 0.650 & 0.344 & 0.386 & 0.855 & 0.393 & 0.738 & 0.344 & 0.539 & 0.858 & 0.225 & 0.686 & 0.704 & 0.234 & 0.392 & 0.349 & \textbf{0.513} \\
\multicolumn{17}{@{}l}{\textit{Qwen3-0.6B retriever, depth.}} \\
6:1024 & 0.488 & 0.260 & 0.314 & 0.804 & 0.248 & 0.603 & 0.279 & 0.405 & 0.835 & 0.168 & 0.580 & 0.646 & 0.219 & 0.330 & 0.284 & \textbf{0.431} \\
10:1024 & 0.554 & 0.292 & 0.330 & 0.811 & 0.289 & 0.609 & 0.314 & 0.447 & 0.844 & 0.186 & 0.605 & 0.739 & 0.257 & 0.341 & 0.283 & \textbf{0.460} \\
14:1024 & 0.592 & 0.321 & 0.326 & 0.817 & 0.320 & 0.621 & 0.317 & 0.493 & 0.845 & 0.196 & 0.620 & 0.767 & 0.237 & 0.350 & 0.289 & \textbf{0.474} \\
18:1024 & 0.634 & 0.312 & 0.338 & 0.820 & 0.336 & 0.619 & 0.320 & 0.492 & 0.838 & 0.199 & 0.616 & 0.753 & 0.210 & 0.355 & 0.310 & \textbf{0.477} \\
24:1024 & 0.641 & 0.293 & 0.368 & 0.837 & 0.353 & 0.673 & 0.312 & 0.509 & 0.845 & 0.203 & 0.639 & 0.699 & 0.230 & 0.368 & 0.314 & \textbf{0.486} \\
28:1024 & 0.661 & 0.339 & 0.389 & 0.856 & 0.387 & 0.734 & 0.346 & 0.550 & 0.854 & 0.214 & 0.683 & 0.735 & 0.268 & 0.393 & 0.359 & \textbf{0.518} \\
\multicolumn{17}{@{}l}{\textit{Qwen3-0.6B retriever, width (MRL).}} \\
28:64 & 0.470 & 0.231 & 0.213 & 0.666 & 0.265 & 0.392 & 0.226 & 0.334 & 0.820 & 0.172 & 0.481 & 0.482 & 0.228 & 0.311 & 0.183 & \textbf{0.365} \\
28:128 & 0.542 & 0.278 & 0.295 & 0.786 & 0.318 & 0.567 & 0.279 & 0.418 & 0.838 & 0.197 & 0.549 & 0.577 & 0.244 & 0.351 & 0.251 & \textbf{0.433} \\
28:256 & 0.579 & 0.301 & 0.338 & 0.816 & 0.351 & 0.644 & 0.306 & 0.460 & 0.846 & 0.207 & 0.597 & 0.625 & 0.257 & 0.368 & 0.287 & \textbf{0.465} \\
28:512 & 0.612 & 0.320 & 0.362 & 0.833 & 0.375 & 0.689 & 0.323 & 0.487 & 0.853 & 0.215 & 0.620 & 0.663 & 0.275 & 0.381 & 0.320 & \textbf{0.489} \\
28:768 & 0.623 & 0.333 & 0.370 & 0.846 & 0.384 & 0.713 & 0.333 & 0.496 & 0.855 & 0.220 & 0.640 & 0.671 & 0.271 & 0.387 & 0.338 & \textbf{0.499} \\
28:1024 & 0.628 & 0.343 & 0.382 & 0.855 & 0.389 & 0.726 & 0.339 & 0.503 & 0.857 & 0.222 & 0.646 & 0.676 & 0.266 & 0.391 & 0.345 & \textbf{0.505} \\
\multicolumn{17}{@{}l}{\textit{Qwen3-0.6B retriever, depth$\times$width (2D Matryoshka).}} \\
6:64 & 0.346 & 0.117 & 0.101 & 0.501 & 0.098 & 0.228 & 0.139 & 0.144 & 0.775 & 0.106 & 0.318 & 0.457 & 0.195 & 0.141 & 0.085 & \textbf{0.250} \\
10:128 & 0.460 & 0.216 & 0.195 & 0.683 & 0.176 & 0.393 & 0.210 & 0.289 & 0.824 & 0.150 & 0.453 & 0.563 & 0.248 & 0.235 & 0.160 & \textbf{0.350} \\
14:256 & 0.565 & 0.301 & 0.287 & 0.777 & 0.272 & 0.527 & 0.291 & 0.440 & 0.842 & 0.191 & 0.564 & 0.669 & 0.246 & 0.317 & 0.237 & \textbf{0.435} \\
18:512 & 0.620 & 0.329 & 0.326 & 0.807 & 0.333 & 0.581 & 0.313 & 0.495 & 0.839 & 0.197 & 0.601 & 0.715 & 0.242 & 0.346 & 0.293 & \textbf{0.469} \\
24:768 & 0.657 & 0.320 & 0.351 & 0.820 & 0.351 & 0.641 & 0.302 & 0.511 & 0.842 & 0.202 & 0.638 & 0.635 & 0.226 & 0.362 & 0.312 & \textbf{0.478} \\
28:1024 & 0.677 & 0.362 & 0.389 & 0.850 & 0.396 & 0.731 & 0.341 & 0.560 & 0.857 & 0.220 & 0.693 & 0.730 & 0.256 & 0.390 & 0.360 & \textbf{0.521} \\
\multicolumn{17}{@{}l}{\textit{Qwen3-0.6B retriever, token (MLTC, pool@20).}} \\
L28xR0.4xP20 & 0.646 & 0.349 & 0.357 & 0.827 & 0.342 & 0.685 & 0.334 & 0.532 & 0.841 & 0.211 & 0.677 & 0.712 & 0.232 & 0.370 & 0.333 & \textbf{0.497} \\
L28xR0.6xP20 & 0.657 & 0.363 & 0.369 & 0.832 & 0.366 & 0.704 & 0.329 & 0.540 & 0.850 & 0.215 & 0.683 & 0.734 & 0.245 & 0.374 & 0.342 & \textbf{0.507} \\
L28xR0.8xP20 & 0.660 & 0.369 & 0.376 & 0.845 & 0.376 & 0.713 & 0.338 & 0.546 & 0.852 & 0.217 & 0.681 & 0.765 & 0.258 & 0.378 & 0.348 & \textbf{0.515} \\
L28xR1.0xP20 & 0.655 & 0.377 & 0.381 & 0.842 & 0.368 & 0.722 & 0.339 & 0.546 & 0.855 & 0.219 & 0.669 & 0.712 & 0.245 & 0.389 & 0.349 & \textbf{0.511} \\
\multicolumn{17}{@{}l}{\textit{Qwen3-0.6B reranker, plain.}} \\
L28 & 0.604 & 0.342 & 0.449 & 0.894 & 0.432 & 0.799 & 0.392 & 0.596 & 0.853 & 0.207 & 0.762 & 0.867 & 0.357 & 0.399 & 0.390 & \textbf{0.556} \\
\multicolumn{17}{@{}l}{\textit{Qwen3-0.6B reranker, depth.}} \\
L4 & 0.011 & 0.047 & 0.062 & 0.111 & 0.009 & 0.027 & 0.130 & 0.016 & 0.006 & 0.011 & 0.011 & 0.307 & 0.102 & 0.015 & 0.005 & \textbf{0.058} \\
L8 & 0.525 & 0.251 & 0.354 & 0.848 & 0.288 & 0.724 & 0.323 & 0.403 & 0.848 & 0.173 & 0.676 & 0.721 & 0.267 & 0.287 & 0.334 & \textbf{0.468} \\
L12 & 0.535 & 0.273 & 0.365 & 0.854 & 0.295 & 0.737 & 0.329 & 0.438 & 0.856 & 0.166 & 0.674 & 0.744 & 0.290 & 0.312 & 0.314 & \textbf{0.479} \\
L16 & 0.593 & 0.340 & 0.456 & 0.890 & 0.424 & 0.796 & 0.382 & 0.591 & 0.860 & 0.201 & 0.758 & 0.870 & 0.361 & 0.399 & 0.396 & \textbf{0.554} \\
L20 & 0.600 & 0.344 & 0.457 & 0.892 & 0.420 & 0.796 & 0.390 & 0.592 & 0.829 & 0.203 & 0.759 & 0.864 & 0.368 & 0.400 & 0.391 & \textbf{0.554} \\
L24 & 0.597 & 0.330 & 0.447 & 0.889 & 0.420 & 0.793 & 0.390 & 0.589 & 0.778 & 0.199 & 0.754 & 0.858 & 0.350 & 0.400 & 0.386 & \textbf{0.545} \\
L28 & 0.603 & 0.337 & 0.453 & 0.891 & 0.422 & 0.797 & 0.389 & 0.592 & 0.848 & 0.201 & 0.755 & 0.862 & 0.368 & 0.400 & 0.389 & \textbf{0.554} \\
\multicolumn{17}{@{}l}{\textit{Qwen3-0.6B reranker, token (LTC; pool@20, exit@28; legacy keys).}} \\
L20xR0.4 & 0.572 & 0.337 & 0.450 & 0.884 & 0.406 & 0.796 & 0.380 & 0.593 & 0.755 & 0.203 & 0.755 & 0.863 & 0.305 & 0.401 & 0.368 & \textbf{0.538} \\
L20xR0.6 & 0.593 & 0.338 & 0.443 & 0.885 & 0.406 & 0.796 & 0.383 & 0.590 & 0.758 & 0.204 & 0.742 & 0.853 & 0.302 & 0.400 & 0.394 & \textbf{0.539} \\
L20xR0.8 & 0.578 & 0.338 & 0.448 & 0.886 & 0.409 & 0.796 & 0.384 & 0.591 & 0.814 & 0.203 & 0.753 & 0.859 & 0.319 & 0.401 & 0.382 & \textbf{0.544} \\
L20xR1.0 & 0.610 & 0.331 & 0.449 & 0.891 & 0.423 & 0.798 & 0.388 & 0.591 & 0.842 & 0.204 & 0.754 & 0.869 & 0.341 & 0.401 & 0.394 & \textbf{0.552} \\
\bottomrule
\end{tabular}
}
\end{table*}

\paragraph{What the experiments assess.} The experiments assess the desiderata discussed in
\cref{sec:framework}. The backbone coverage exercises the common interface across encoder and
decoder families. The quality curves test whether a common operating-point description and schedule
can express the expected depth, width, token, and composed trade-offs. The experiments also measure
the cost of joint training at the full point (the \emph{elasticity tax}: elastic full-point minus
the dedicated plain full-point), and separately measure the deployment costs that each axis is
intended to change. We focus less on absolute performance numbers and more on diagnosing the
framework. Every number we report comes from a single training run per configuration; we do not repeat runs with different seeds and so cannot report variance. We accordingly draw conclusions from the shape of each curve, which is consistent across the three backbones for the depth, width, and depth$\times$width axes (the token axis is evaluated on Qwen3-0.6B only), rather than from individual small differences, and we flag below the comparisons whose margins may lie within run-to-run variation, which we do not measure.

\begin{figure*}[t]
\centering
\input{technical-report/figures/flops-camera-ready.tex}
\caption{Quality versus relative analytic cost. \textbf{Left:} retrieval document encoding, depth
and MLTC, against relative FLOPs; MRL is absent because it does not change FLOPs. \textbf{Center:}
retrieval against relative index size; only MRL moves along this axis. \textbf{Right:} reranking
against relative score FLOPs. Depth and token reduce compute and leave the index unchanged; width
reduces the index and leaves compute unchanged.}
\Description{Three panels compare retrieval document-encoding cost, retrieval index storage, and reranking compute cost against MS MARCO quality.}
\label{fig:eff-flops}
\end{figure*}

\begin{figure*}[t]
\centering
\input{technical-report/figures/speed-camera-ready.tex}
\caption{Quality retained versus measured speedup. \textbf{Left:} retrieval document throughput.
\textbf{Center:} online query latency at batch 1. \textbf{Right:} reranking throughput. Depth speeds
up every path; the token axis helps document encoding modestly and online query encoding not at all,
because these short, padded queries offer little compute to save and the lower layers still run.}
\Description{Three panels compare retained ranking quality against measured document-encoding, query-encoding, and reranking speedups. Depth provides the largest speedups while token compression primarily benefits long inputs.}
\label{fig:eff-speed}
\end{figure*}

\subsection{Results and analysis}

We first evaluate the single-axis operating points and then their compositions. These configurations
correspond to prior elastic methods where applicable. \cref{fig:retr-axes,fig:token,fig:rerank-depth}
show the quality curves (BEIR-15 nDCG@10) while \cref{tab:ds-qwen-all} reports complete results for the Qwen3-0.6B model. We skip the detailed BERT and ModernBERT results for the sake of space. 
Dataset abbreviations: ARG arguana,
CFE climate-fever, DBP dbpedia-entity, FEV fever, FIQ fiqa, HOP hotpotqa, NFC nfcorpus, NQ nq, QUO
quora, SCD scidocs, SCF scifact, TRC trec-covid, TOU webis-touche2020, MSM MS~MARCO dev, CQA
cqadupstack (mean of 12). Granularity labels: $\ell{:}d$ exit layer $\ell$ and dim $d$ (retriever); $\mathrm{L}\ell$
exit layer $\ell$ (reranker); $\mathrm{L}\ell{\times}\mathrm{R}r{\times}\mathrm{P}p$ exit layer $\ell$, keep-ratio $r$, pool
layer $p$ (token / MLTC). For the reranker's legacy token keys $\mathrm{L}p{\times}\mathrm{R}r$, $p$ is the pooling layer, and the exit is the final layer (28 here).

\paragraph{Depth.} For the retriever, performance improves smoothly with the number of layers on all three
backbones (\cref{fig:retr-axes}a; Qwen3 goes from $0.431$ at layer 6 to $0.518$ at layer 28, full
width). The reranker has a different curve (\cref{fig:rerank-depth}): quality is very low when only
a few layers run, then rises and levels off at about half the layers (Qwen3 reaches $0.554$ by layer
16, matching layer 28; BERT levels off at layer 6, ModernBERT at layer 16). The retriever, which
reads out a pooled vector, degrades gradually; the reranker, which reads out a single score, has a
low-quality region at shallow depth and a plateau at greater depth. On both tasks, ModernBERT shows a step near layer 13 rather than a gradual rise.

\paragraph{Width (MRL)} The pure-width curve is smooth and monotonic on all three backbones
(\cref{fig:retr-axes}b; e.g., Qwen3 $0.365 \to 0.505$ from dim 64 to 1024 at full depth). This
reproduces MRL as a configuration; the corresponding cost is index size rather than compute
(\cref{sec:efficiency}).

\paragraph{Token (LTC Reranker and MLTC Retriever)} On the reranker, we follow \citet{Zhuang2026LayerwiseTC} to pool the sequence at layer 20 and sweep the
keep-ratio for multiple runs; quality falls gently and never collapses (\cref{fig:token}; Qwen3 $0.552$ at ratio 1.0
to $0.538$ at ratio 0.4). We then apply the same pooling operation to the \emph{retriever}\,---\,the
identical code, now on a bi-encoder\,---\,and obtain a curve with a similarly gentle shape ($0.497$ to $0.511$). Different from the LTC reranker which trains a single (layer, ratio) point, we use MLTC, which jointly trains several ratios in one checkpoint. At keep-ratio 1.0, the MLTC retriever's full-point quality is close to that of the plain Qwen3-0.6B retriever (0.511 vs 0.513). We do not compare it directly with separately trained single-ratio LTC retrievers, so this experiment does not quantify the benefit of joint training on the token axis, the token-axis counterpart of the elasticity tax in \cref{tab:tax}; we leave this controlled comparison to future work.

\paragraph{Composition: depth $\times$ width.} The 2D~Matryoshka schedule recovers the Starbucks-style frontier on
all three backbones (\cref{fig:retr-axes}c); each frontier is smooth and monotonic (BERT
$0.263 \to 0.448$; Qwen3 $0.250 \to 0.521$; ModernBERT $0.187 \to 0.450$ across six operating points
from one checkpoint).
The shape matches Starbucks. This is a comparison of curve shape rather than of numbers.
\par\newpage

The released Starbucks checkpoints are trained with masked-autoencoding pretraining and on different data, neither of which is part of our recipe, and the same holds for released MRL and LTC checkpoints. Because the recipes differ, comparing against those checkpoints on BEIR-15 would measure the training recipe as much as the abstraction, which is why our plain single-point baselines, trained with the identical recipe, serve as the controlled reference at the full operating point (\cref{tab:tax}). The released-checkpoint evaluation, and a comparison that also controls training data at compressed operating points, are left to future work.

\paragraph{Full-point elasticity tax.} Training elastic models may degrade model performance compared to standard training; we refer to this cost as the \emph{elasticity tax}. \cref{tab:tax} compares the full model trained in the elastic way to a
model trained for that size alone, using the depth$\times$width schedule
so the comparison is the same across backbones. We note the measured difference is small: on the retriever it is
$-0.013$ and $-0.026$ for the two encoders and $+0.008$ for the decoder. The full point
changes modestly on the two encoder backbones, and the Qwen3
decoder has the smallest tax. Each entry is a single run, so the sign of the smallest differences, such as the $+0.008$ for Qwen3, should not be over-interpreted; the reading we take from these runs is that the tax is small on every backbone.

\begin{table}[t]
\centering
\small
\caption{Elasticity tax at the full operating point (BEIR-15 nDCG@10). The depth$\times$width schedule is used for all backbones for comparability.}
\label{tab:tax}
\begin{tabular}{lccc}
\toprule
& plain & elastic (2D) & tax \\
\midrule
BERT @ 12:768       & 0.461 & 0.448 & $-0.013$ \\
ModernBERT @ 22:768 & 0.476 & 0.450 & $-0.026$ \\
Qwen3 @ 28:1024     & 0.513 & 0.521 & $+0.008$ \\
\bottomrule
\end{tabular}
\end{table}

\begin{samepage}
\paragraph{Which axes each task admits.} A retriever produces a vector and admits all three axes; a
reranker produces a scalar score and admits depth and token but not width, since there is no
embedding to truncate. This is the one structural exclusion in the task-by-axis matrix, explaining the reranker's smaller configuration set.
\par\end{samepage}

\paragraph{The three-axis composition is left open.} A depth $\times$ width $\times$ token retriever
is expressible today (e.g., \texttt{"24:256@0.6/20"}), and the machinery runs it (training and
inference both route through the \texttt{pool\_layer} branch of \cref{sec:framework}). We do not
report an operating point for it: this paper verifies only that the machinery runs, and the quality--cost trade-off of the three-axis composition is not evaluated here; we leave this experiment to future work.

\flushbottom
\vspace{10pt}
\subsection{Efficiency and deployment trade-offs}
\label{sec:efficiency}

We measure deployment costs for Qwen3-0.6B on MS~MARCO dev, using MRR@10 for quality.
A single in-domain corpus avoids BEIR's domain and length confounds. Timing uses one GPU and a fixed workload, separately from parallel evaluation.
We report document encoding (offline, batch 128, length 512, documents/second) and query encoding
(request path, batch 1, latency/query). Queries are padded to 32 tokens to cover long inputs, although they average about 8 tokens.
For reranking, we host the model through a Hugging Face backend, send HTTP requests in batches of 128 query--document
pairs, and compute throughput from the end-to-end request latency.
\cref{fig:eff-flops,fig:eff-speed} show the measured costs and quality trade-offs on both tasks.

\paragraph{The measured speedup follows the cost model.} For reranking at batch size 128, the measured
speedup agrees with the prediction ($1/\text{relative FLOPs}$) to within about two percent at every
depth (layer 4 measures $6.86\times$ against a predicted $7.00\times$; layer 16, $1.75\times$
against $1.75\times$). Document encoding tracks just as closely (layer 6, $4.61\times$ measured
against $4.67\times$ predicted). When the batch is large the work is dominated by arithmetic, so
counting FLOPs is enough to predict the wall-clock speedup.

\paragraph{Depth and token help different paths.} At layer 16, the reranker retains the full-depth
quality ($0.342$ MRR@10) while running $1.75\times$ faster. The token axis gives a smaller speedup because
the layers below the pooling point still run on the full sequence (pooling at layer 20 of 28 leaves
about $0.71$ of the work in place), though its quality falls more gently. The gap is starker for
queries: cutting depth still speeds up query encoding by about $4\times$ at layer 6, while token
pooling gives no speedup (around $0.99\times$, slightly slower), since these short queries, padded to 32 tokens in the timing study, offer little work to save
and the lower layers run regardless. Token compression is therefore a lever for long documents,
while depth speeds up both the offline corpus pass and the online query. At batch 1 the query pass
spends much of its time on overhead rather than arithmetic, so its speedup falls short of the FLOP
prediction\,---\,which is why we measure query latency directly instead of reading it off the FLOP
count for these short inputs.

\paragraph{Width reduces storage and search.} With an exact inner-product index, both the stored
index size and the search FLOPs are proportional to the embedding dimension, so index size measures
the width axis's cost independently of hardware. Reducing the Qwen3 embedding from dim 1024 to
dim 64 reduces the MS~MARCO index from about 36\,GB to about 2.3\,GB while MRR@10 moves from 0.329 to
0.261; depth and token leave the index size unchanged.

\paragraph{Depth and token save compute; width saves storage.} Taken together, the three axes address different
deployment constraints: depth reduces compute on every path and is the most broadly applicable axis;
the token axis reduces compute on long inputs specifically (corpus encoding, not short queries); and
width reduces index storage and search cost. The cost model and the figures above indicate which
axis to reach for given a latency or storage budget.

These deployment measurements concern inference rather than training overhead. Depth and width points reuse the backbone pass, whereas each token point adds an upper-layer pass on its pooled sequence. Serving several operating points from one checkpoint therefore does not imply equal training costs for every schedule; timing and memory remain separate evaluation questions.

\vspace{3pt}
\section{Conclusion and Future Work}
\label{sec:conclusion}

Existing model compression work in the IR community mostly targets a single compression axis at a time. 
We challenge this \emph{status quo} by first providing a taxonomy of three compression axes, then proposing \ours to study the three axes in a unified interface.
Through one small dataclass and a declarative schedule, \ours expresses depth and token compression for both tasks and width compression for retrieval. It expresses four prior methods\,---\,MRL,
early exit, 2D~Matryoshka, and LTC\,---\,as configurations, and runs across five encoder and decoder
backbone families without rewriting their model classes (the token axis adds one small per-family shim). Across 20
checkpoints, we show that the elastic curves are smooth with few exceptions, full-point quality remains close to that of dedicated single-point models, and the measured inference speedup follows the analytic cost model. The same interface enables a new algorithm MLTC, which serves several token-compression ratios from one retriever checkpoint. We release the code and all checkpoints for community use.

Our experience building~\ours suggests several directions for extending the framework and evaluating its trade-offs. First, the token axis currently targets global-attention backbones; extending it to
sliding-window-attention models (such as ModernBERT) requires a pooling rule compatible with a fixed
local window, after which the token axis applies to that family as well. Second, the abstraction
already expresses the full depth$\times$width$\times$token composition on the retriever (\cref{sec:experiments}); a
practitioner with a specific latency and storage budget can train and select that operating point
directly, although we have not evaluated this composition and its quality--cost frontier remains to be measured. 
Third, the empirical study has some limits that the released code makes straightforward to close: for example, MLTC is not compared against separately trained single-ratio LTC retrievers; and training time and peak memory as a function of schedule size are not reported. Because a new axis, backbone, or per-granularity loss enters through a single interface,
we expect further compression methods to slot in as additional configurations rather than as new systems.  Finally, elastic modeling for learned sparse retrieval~\citep{formal2021spladev1,formal2022spladeplus,xu2025cspladelearnedsparseretrieval,xu2026laconic} remains underexplored.

\clearpage
\newcounter{balancebib}
\let\oldbibitem\bibitem
\def\bibitem{\stepcounter{balancebib}\ifnum\value{balancebib}=46 \vfill\eject\fi\oldbibitem}

\bibliographystyle{ACM-Reference-Format}

\bibliography{technical-report/references,acmart-primary/camera-ready-additions}

\end{document}

%% file: technical-report/figures/flops-camera-ready.tex
\begin{tikzpicture}
  \node[anchor=south west,inner sep=0] (plot) at (0,0)
    {\includegraphics[width=\textwidth]{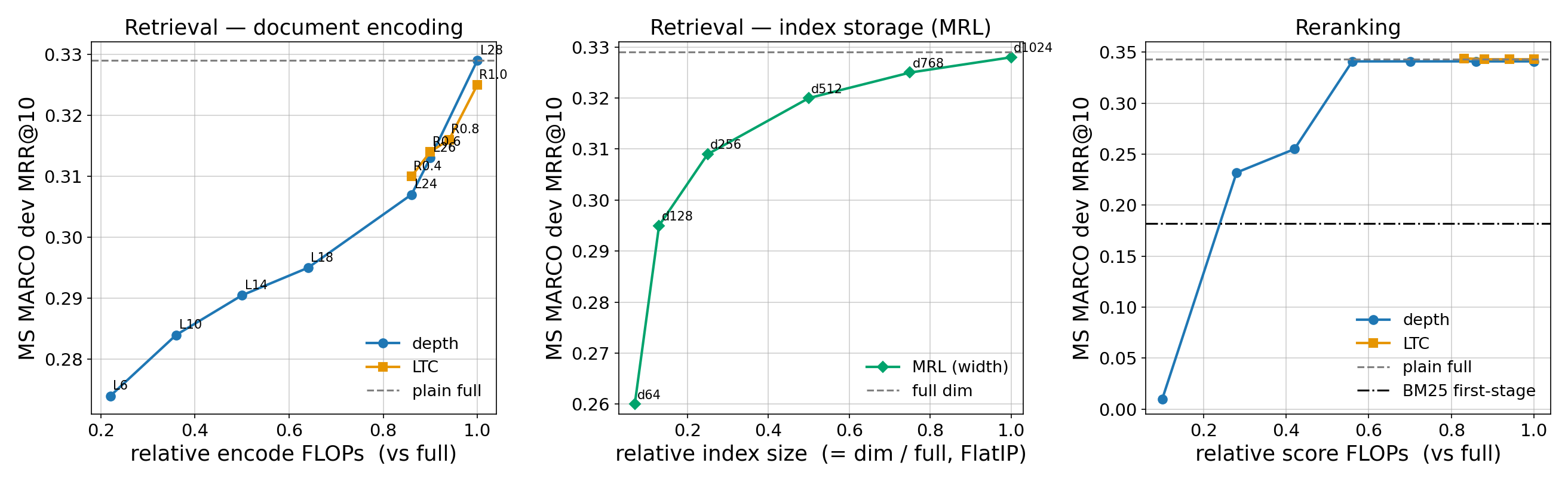}};
  \begin{scope}[x={(plot.south east)},y={(plot.north west)}]
    \fill[white] (0.261,0.224) rectangle (0.291,0.261);
    \node[anchor=base west,inner sep=0,font=\fontfamily{phv}\fontsize{6.0}{6.5}\selectfont]
      at (0.262,0.230) {MLTC};
  \end{scope}
\end{tikzpicture}%

%% file: technical-report/figures/speed-camera-ready.tex
\begin{tikzpicture}
  \node[anchor=south west,inner sep=0] (plot) at (0,0)
    {\includegraphics[width=\textwidth]{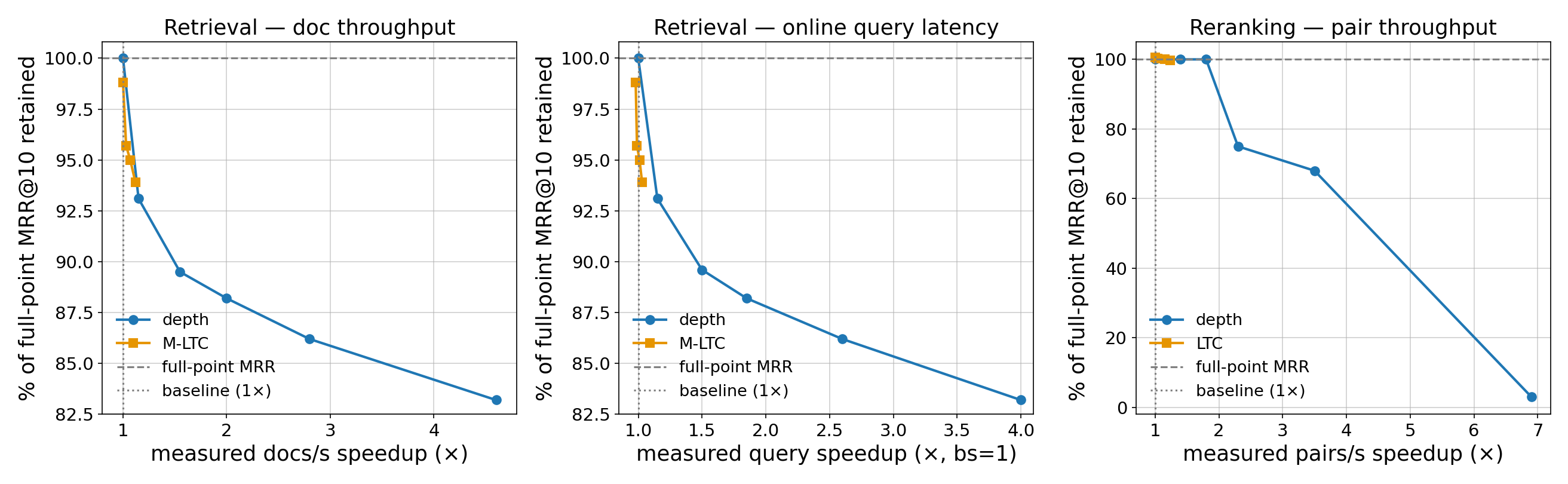}};
  \begin{scope}[x={(plot.south east)},y={(plot.north west)}]
    \fill[white] (0.103,0.272) rectangle (0.134,0.306);
    \fill[white] (0.433,0.272) rectangle (0.464,0.306);
    \node[anchor=base west,inner sep=0,font=\fontfamily{phv}\fontsize{6.0}{6.5}\selectfont]
      at (0.1035,0.279) {MLTC};
    \node[anchor=base west,inner sep=0,font=\fontfamily{phv}\fontsize{6.0}{6.5}\selectfont]
      at (0.4335,0.279) {MLTC};
  \end{scope}
\end{tikzpicture}%

%% file: acm_v2_19/acmmanuscript.bbl

\begin{thebibliography}{54}


\ifx \showCODEN    \undefined \def \showCODEN     #1{\unskip}     \fi
\ifx \showISBNx    \undefined \def \showISBNx     #1{\unskip}     \fi
\ifx \showISBNxiii \undefined \def \showISBNxiii  #1{\unskip}     \fi
\ifx \showISSN     \undefined \def \showISSN      #1{\unskip}     \fi
\ifx \showLCCN     \undefined \def \showLCCN      #1{\unskip}     \fi
\ifx \shownote     \undefined \def \shownote      #1{#1}          \fi
\ifx \showarticletitle \undefined \def \showarticletitle #1{#1}   \fi
\ifx \showURL      \undefined \def \showURL       {\relax}        \fi
\providecommand\bibfield[2]{#2}
\providecommand\bibinfo[2]{#2}
\providecommand\natexlab[1]{#1}
\providecommand\showeprint[2][]{arXiv:#2}

\bibitem[Bolya et~al\mbox{.}(2023)]%
        {bolya2023tome}
\bibfield{author}{\bibinfo{person}{Daniel Bolya}, \bibinfo{person}{Cheng-Yang Fu}, \bibinfo{person}{Xiaoliang Dai}, \bibinfo{person}{Peizhao Zhang}, \bibinfo{person}{Christoph Feichtenhofer}, {and} \bibinfo{person}{Judy Hoffman}.} \bibinfo{year}{2023}\natexlab{}.
\newblock \showarticletitle{Token Merging: Your {ViT} But Faster}. In \bibinfo{booktitle}{\emph{International Conference on Learning Representations (ICLR)}}.
\newblock


\bibitem[Cai et~al\mbox{.}(2020)]%
        {cai2020onceforall}
\bibfield{author}{\bibinfo{person}{Han Cai}, \bibinfo{person}{Chuang Gan}, \bibinfo{person}{Tianzhe Wang}, \bibinfo{person}{Zhekai Zhang}, {and} \bibinfo{person}{Song Han}.} \bibinfo{year}{2020}\natexlab{}.
\newblock \showarticletitle{Once-for-All: Train One Network and Specialize It for Efficient Deployment}. In \bibinfo{booktitle}{\emph{International Conference on Learning Representations (ICLR)}}.
\newblock


\bibitem[Cai et~al\mbox{.}(2024)]%
        {cai2024flextron}
\bibfield{author}{\bibinfo{person}{Ruisi Cai}, \bibinfo{person}{Saurav Muralidharan}, \bibinfo{person}{Greg Heinrich}, \bibinfo{person}{Hongxu Yin}, \bibinfo{person}{Zhangyang Wang}, \bibinfo{person}{Jan Kautz}, {and} \bibinfo{person}{Pavlo Molchanov}.} \bibinfo{year}{2024}\natexlab{}.
\newblock \showarticletitle{Flextron: Many-in-One Flexible Large Language Model}. In \bibinfo{booktitle}{\emph{Proceedings of the 41st International Conference on Machine Learning (ICML)}}.
\newblock


\bibitem[Cao et~al\mbox{.}(2007)]%
        {cao2007learningtorank}
\bibfield{author}{\bibinfo{person}{Zhe Cao}, \bibinfo{person}{Tao Qin}, \bibinfo{person}{Tie-Yan Liu}, \bibinfo{person}{Ming-Feng Tsai}, {and} \bibinfo{person}{Hang Li}.} \bibinfo{year}{2007}\natexlab{}.
\newblock \showarticletitle{Learning to Rank: From Pairwise Approach to Listwise Approach}. In \bibinfo{booktitle}{\emph{Proceedings of the 24th International Conference on Machine Learning}}. \bibinfo{pages}{129--136}.
\newblock


\bibitem[Dettmers et~al\mbox{.}(2022)]%
        {dettmers2022llmint8}
\bibfield{author}{\bibinfo{person}{Tim Dettmers}, \bibinfo{person}{Mike Lewis}, \bibinfo{person}{Younes Belkada}, {and} \bibinfo{person}{Luke Zettlemoyer}.} \bibinfo{year}{2022}\natexlab{}.
\newblock \showarticletitle{{LLM.int8()}: 8-bit Matrix Multiplication for Transformers at Scale}.
\newblock \bibinfo{journal}{\emph{Advances in Neural Information Processing Systems}}  \bibinfo{volume}{35} (\bibinfo{year}{2022}), \bibinfo{pages}{30318--30332}.
\newblock


\bibitem[Devlin et~al\mbox{.}(2019)]%
        {devlin-etal-2019-bert}
\bibfield{author}{\bibinfo{person}{Jacob Devlin}, \bibinfo{person}{Ming-Wei Chang}, \bibinfo{person}{Kenton Lee}, {and} \bibinfo{person}{Kristina Toutanova}.} \bibinfo{year}{2019}\natexlab{}.
\newblock \showarticletitle{{BERT}: Pre-training of Deep Bidirectional Transformers for Language Understanding}. In \bibinfo{booktitle}{\emph{Proceedings of the 2019 Conference of the North {A}merican Chapter of the Association for Computational Linguistics: Human Language Technologies, Volume 1 (Long and Short Papers)}}, \bibfield{editor}{\bibinfo{person}{Jill Burstein}, \bibinfo{person}{Christy Doran}, {and} \bibinfo{person}{Thamar Solorio}} (Eds.). \bibinfo{publisher}{Association for Computational Linguistics}, \bibinfo{address}{Minneapolis, Minnesota}, \bibinfo{pages}{4171--4186}.
\newblock
\href{https://doi.org/10.18653/v1/N19-1423}{doi:\nolinkurl{10.18653/v1/N19-1423}}


\bibitem[Devvrit et~al\mbox{.}(2023)]%
        {devvrit2023matformer}
\bibfield{author}{\bibinfo{person}{Devvrit}, \bibinfo{person}{Sneha Kudugunta}, \bibinfo{person}{Aditya Kusupati}, \bibinfo{person}{Tim Dettmers}, \bibinfo{person}{Kaifeng Chen}, \bibinfo{person}{Inderjit Dhillon}, \bibinfo{person}{Yulia Tsvetkov}, \bibinfo{person}{Hannaneh Hajishirzi}, \bibinfo{person}{Sham Kakade}, \bibinfo{person}{Ali Farhadi}, {and} \bibinfo{person}{Prateek Jain}.} \bibinfo{year}{2023}\natexlab{}.
\newblock \bibinfo{title}{{MatFormer}: Nested Transformer for Elastic Inference}.
\newblock
\showeprint[arxiv]{2310.07707}~[cs.LG]
\urldef\tempurl%
\url{https://arxiv.org/abs/2310.07707}
\showURL{%
\tempurl}


\bibitem[Elhage et~al\mbox{.}(2021)]%
        {elhage2021mathematicalframework}
\bibfield{author}{\bibinfo{person}{Nelson Elhage}, \bibinfo{person}{Neel Nanda}, \bibinfo{person}{Catherine Olsson}, \bibinfo{person}{Tom Henighan}, \bibinfo{person}{Nicholas Joseph}, \bibinfo{person}{Ben Mann}, \bibinfo{person}{Amanda Askell}, \bibinfo{person}{Yuntao Bai}, \bibinfo{person}{Anna Chen}, \bibinfo{person}{Tom Conerly}, \bibinfo{person}{Nova DasSarma}, \bibinfo{person}{Dawn Drain}, \bibinfo{person}{Deep Ganguli}, \bibinfo{person}{Zac Hatfield-Dodds}, \bibinfo{person}{Danny Hernandez}, \bibinfo{person}{Andy Jones}, \bibinfo{person}{Jackson Kernion}, \bibinfo{person}{Liane Lovitt}, \bibinfo{person}{Kamal Ndousse}, \bibinfo{person}{Dario Amodei}, \bibinfo{person}{Tom Brown}, \bibinfo{person}{Jack Clark}, \bibinfo{person}{Jared Kaplan}, \bibinfo{person}{Sam McCandlish}, {and} \bibinfo{person}{Chris Olah}.} \bibinfo{year}{2021}\natexlab{}.
\newblock \showarticletitle{A Mathematical Framework for Transformer Circuits}.
\newblock \bibinfo{journal}{\emph{Transformer Circuits Thread}} (\bibinfo{year}{2021}).
\newblock
\shownote{https://transformer-circuits.pub/2021/framework/index.html}.
\newblock


\bibitem[Elhoushi et~al\mbox{.}(2024)]%
        {Elhoushi2024layerskip}
\bibfield{author}{\bibinfo{person}{Mostafa Elhoushi}, \bibinfo{person}{Akshat Shrivastava}, \bibinfo{person}{Diana Liskovich}, \bibinfo{person}{Basil Hosmer}, \bibinfo{person}{Bram Wasti}, \bibinfo{person}{Liangzhen Lai}, \bibinfo{person}{Anas Mahmoud}, \bibinfo{person}{Bilge Acun}, \bibinfo{person}{Saurabh Agarwal}, \bibinfo{person}{Ahmed Roman}, \bibinfo{person}{Ahmed~A Aly}, \bibinfo{person}{Beidi Chen}, {and} \bibinfo{person}{Carole-Jean Wu}.} \bibinfo{year}{2024}\natexlab{}.
\newblock \showarticletitle{LayerSkip: Enabling Early Exit Inference and Self-Speculative Decoding}. In \bibinfo{booktitle}{\emph{Proceedings of the 62nd Annual Meeting of the Association for Computational Linguistics (Volume 1: Long Papers)}}. \bibinfo{publisher}{Association for Computational Linguistics}, \bibinfo{address}{Bangkok, Thailand}, \bibinfo{pages}{12622--12642}.
\newblock
\href{https://doi.org/10.18653/v1/2024.acl-long.681}{doi:\nolinkurl{10.18653/v1/2024.acl-long.681}}


\bibitem[Fan et~al\mbox{.}(2020)]%
        {fan2020layerdrop}
\bibfield{author}{\bibinfo{person}{Angela Fan}, \bibinfo{person}{Edouard Grave}, {and} \bibinfo{person}{Armand Joulin}.} \bibinfo{year}{2020}\natexlab{}.
\newblock \showarticletitle{Reducing Transformer Depth on Demand with Structured Dropout}. In \bibinfo{booktitle}{\emph{International Conference on Learning Representations (ICLR)}}.
\newblock


\bibitem[Formal et~al\mbox{.}(2022)]%
        {formal2022spladeplus}
\bibfield{author}{\bibinfo{person}{Thibault Formal}, \bibinfo{person}{Carlos Lassance}, \bibinfo{person}{Benjamin Piwowarski}, {and} \bibinfo{person}{St\'{e}phane Clinchant}.} \bibinfo{year}{2022}\natexlab{}.
\newblock \showarticletitle{From Distillation to Hard Negative Sampling: Making Sparse Neural IR Models More Effective}. In \bibinfo{booktitle}{\emph{Proceedings of the 45th International ACM SIGIR Conference on Research and Development in Information Retrieval}} (Madrid, Spain) \emph{(\bibinfo{series}{SIGIR '22})}. \bibinfo{publisher}{Association for Computing Machinery}, \bibinfo{address}{New York, NY, USA}, \bibinfo{pages}{2353–2359}.
\newblock
\showISBNx{9781450387323}
\href{https://doi.org/10.1145/3477495.3531857}{doi:\nolinkurl{10.1145/3477495.3531857}}


\bibitem[Formal et~al\mbox{.}(2021)]%
        {formal2021spladev1}
\bibfield{author}{\bibinfo{person}{Thibault Formal}, \bibinfo{person}{Benjamin Piwowarski}, {and} \bibinfo{person}{St\'{e}phane Clinchant}.} \bibinfo{year}{2021}\natexlab{}.
\newblock \showarticletitle{SPLADE: Sparse Lexical and Expansion Model for First Stage Ranking}. In \bibinfo{booktitle}{\emph{Proceedings of the 44th International ACM SIGIR Conference on Research and Development in Information Retrieval}}. \bibinfo{publisher}{Association for Computing Machinery}, \bibinfo{address}{New York, NY, USA}, \bibinfo{pages}{2288–2292}.
\newblock
\showISBNx{9781450380379}
\urldef\tempurl%
\url{https://doi.org/10.1145/3404835.3463098}
\showURL{%
\tempurl}


\bibitem[Frantar et~al\mbox{.}(2023)]%
        {frantar2023gptq}
\bibfield{author}{\bibinfo{person}{Elias Frantar}, \bibinfo{person}{Saleh Ashkboos}, \bibinfo{person}{Torsten Hoefler}, {and} \bibinfo{person}{Dan Alistarh}.} \bibinfo{year}{2023}\natexlab{}.
\newblock \showarticletitle{{OPTQ}: Accurate Quantization for Generative Pre-trained Transformers}. In \bibinfo{booktitle}{\emph{The Eleventh International Conference on Learning Representations}}.
\newblock
\urldef\tempurl%
\url{https://openreview.net/forum?id=tcbBPnfwxS}
\showURL{%
\tempurl}


\bibitem[Gao and Callan(2022)]%
        {gao-callan-2022-cocondenser}
\bibfield{author}{\bibinfo{person}{Luyu Gao} {and} \bibinfo{person}{Jamie Callan}.} \bibinfo{year}{2022}\natexlab{}.
\newblock \showarticletitle{Unsupervised Corpus Aware Language Model Pre-training for Dense Passage Retrieval}. In \bibinfo{booktitle}{\emph{Proceedings of the 60th Annual Meeting of the Association for Computational Linguistics (Volume 1: Long Papers)}}, \bibfield{editor}{\bibinfo{person}{Smaranda Muresan}, \bibinfo{person}{Preslav Nakov}, {and} \bibinfo{person}{Aline Villavicencio}} (Eds.). \bibinfo{publisher}{Association for Computational Linguistics}, \bibinfo{address}{Dublin, Ireland}, \bibinfo{pages}{2843--2853}.
\newblock
\href{https://doi.org/10.18653/v1/2022.acl-long.203}{doi:\nolinkurl{10.18653/v1/2022.acl-long.203}}


\bibitem[Goyal et~al\mbox{.}(2020)]%
        {goyal2020powerbert}
\bibfield{author}{\bibinfo{person}{Saurabh Goyal}, \bibinfo{person}{Anamitra~Roy Choudhury}, \bibinfo{person}{Saurabh Raje}, \bibinfo{person}{Venkatesan Chakaravarthy}, \bibinfo{person}{Yogish Sabharwal}, {and} \bibinfo{person}{Ashish Verma}.} \bibinfo{year}{2020}\natexlab{}.
\newblock \showarticletitle{{P}o{WER}-{BERT}: Accelerating {BERT} Inference via Progressive Word-vector Elimination}. In \bibinfo{booktitle}{\emph{Proceedings of the 37th International Conference on Machine Learning}} \emph{(\bibinfo{series}{Proceedings of Machine Learning Research}, Vol.~\bibinfo{volume}{119})}, \bibfield{editor}{\bibinfo{person}{Hal~Daumé III} {and} \bibinfo{person}{Aarti Singh}} (Eds.). \bibinfo{publisher}{PMLR}, \bibinfo{pages}{3690--3699}.
\newblock
\urldef\tempurl%
\url{https://proceedings.mlr.press/v119/goyal20a.html}
\showURL{%
\tempurl}


\bibitem[Grattafiori et~al\mbox{.}(2024)]%
        {grattafiori2024llama3technicalreport}
\bibfield{author}{\bibinfo{person}{Aaron Grattafiori}, \bibinfo{person}{Abhimanyu Dubey}, \bibinfo{person}{Abhinav Jauhri}, \bibinfo{person}{Abhinav Pandey}, \bibinfo{person}{Abhishek Kadian}, \bibinfo{person}{Ahmad Al-Dahle}, \bibinfo{person}{Aiesha Letman}, \bibinfo{person}{Akhil Mathur}, \bibinfo{person}{Alan Schelten}, \bibinfo{person}{Alex Vaughan}, {et~al\mbox{.}}} \bibinfo{year}{2024}\natexlab{}.
\newblock \bibinfo{title}{The {Llama} 3 Herd of Models}.
\newblock
\showeprint[arxiv]{2407.21783}~[cs.AI]
\urldef\tempurl%
\url{https://arxiv.org/abs/2407.21783}
\showURL{%
\tempurl}


\bibitem[Hinton et~al\mbox{.}(2015)]%
        {hinton2015distillingknowledgeneuralnetwork}
\bibfield{author}{\bibinfo{person}{Geoffrey Hinton}, \bibinfo{person}{Oriol Vinyals}, {and} \bibinfo{person}{Jeff Dean}.} \bibinfo{year}{2015}\natexlab{}.
\newblock \bibinfo{title}{Distilling the Knowledge in a Neural Network}.
\newblock
\showeprint[arxiv]{1503.02531}~[stat.ML]
\urldef\tempurl%
\url{https://arxiv.org/abs/1503.02531}
\showURL{%
\tempurl}


\bibitem[Hofst{\"a}tter et~al\mbox{.}(2020)]%
        {hofstatter2020marginmse}
\bibfield{author}{\bibinfo{person}{Sebastian Hofst{\"a}tter}, \bibinfo{person}{Sophia Althammer}, \bibinfo{person}{Michael Schr{\"o}der}, \bibinfo{person}{Mete Sertkan}, {and} \bibinfo{person}{Allan Hanbury}.} \bibinfo{year}{2020}\natexlab{}.
\newblock \bibinfo{title}{Improving Efficient Neural Ranking Models with Cross-Architecture Knowledge Distillation}.
\newblock
\showeprint[arxiv]{2010.02666}~[cs.IR]
\urldef\tempurl%
\url{https://arxiv.org/abs/2010.02666}
\showURL{%
\tempurl}


\bibitem[Hou et~al\mbox{.}(2020)]%
        {hou2020dynabert}
\bibfield{author}{\bibinfo{person}{Lu Hou}, \bibinfo{person}{Zhiqi Huang}, \bibinfo{person}{Lifeng Shang}, \bibinfo{person}{Xin Jiang}, \bibinfo{person}{Xiao Chen}, {and} \bibinfo{person}{Qun Liu}.} \bibinfo{year}{2020}\natexlab{}.
\newblock \showarticletitle{{DynaBERT}: Dynamic {BERT} with Adaptive Width and Depth}. In \bibinfo{booktitle}{\emph{Advances in Neural Information Processing Systems}}, Vol.~\bibinfo{volume}{33}.
\newblock


\bibitem[Jiang et~al\mbox{.}(2023)]%
        {Jiang2023Mistral7B}
\bibfield{author}{\bibinfo{person}{Albert~Qiaochu Jiang}, \bibinfo{person}{Alexandre Sablayrolles}, \bibinfo{person}{Arthur Mensch}, \bibinfo{person}{Chris Bamford}, \bibinfo{person}{Devendra~Singh Chaplot}, \bibinfo{person}{Diego de Las~Casas}, \bibinfo{person}{Florian Bressand}, \bibinfo{person}{Gianna Lengyel}, \bibinfo{person}{Guillaume Lample}, \bibinfo{person}{Lucile Saulnier}, \bibinfo{person}{L{\'e}lio~Renard Lavaud}, \bibinfo{person}{Marie-Anne Lachaux}, \bibinfo{person}{Pierre Stock}, \bibinfo{person}{Teven~Le Scao}, \bibinfo{person}{Thibaut Lavril}, \bibinfo{person}{Thomas Wang}, \bibinfo{person}{Timoth{\'e}e Lacroix}, {and} \bibinfo{person}{William~El Sayed}.} \bibinfo{year}{2023}\natexlab{}.
\newblock \bibinfo{title}{{Mistral 7B}}.
\newblock
\showeprint[arxiv]{2310.06825}~[cs.CL]
\urldef\tempurl%
\url{https://arxiv.org/abs/2310.06825}
\showURL{%
\tempurl}


\bibitem[Johnson et~al\mbox{.}(2017)]%
        {johnson2017billionscalesimilaritysearchgpus}
\bibfield{author}{\bibinfo{person}{Jeff Johnson}, \bibinfo{person}{Matthijs Douze}, {and} \bibinfo{person}{Hervé Jégou}.} \bibinfo{year}{2017}\natexlab{}.
\newblock \bibinfo{title}{Billion-Scale Similarity Search with {GPUs}}.
\newblock
\showeprint[arxiv]{1702.08734}~[cs.CV]
\urldef\tempurl%
\url{https://arxiv.org/abs/1702.08734}
\showURL{%
\tempurl}


\bibitem[Kusupati et~al\mbox{.}(2022)]%
        {kusupati2022matryoshka}
\bibfield{author}{\bibinfo{person}{Aditya Kusupati}, \bibinfo{person}{Gantavya Bhatt}, \bibinfo{person}{Aniket Rege}, \bibinfo{person}{Matthew Wallingford}, \bibinfo{person}{Aditya Sinha}, \bibinfo{person}{Vivek Ramanujan}, \bibinfo{person}{William Howard-Snyder}, \bibinfo{person}{Kaifeng Chen}, \bibinfo{person}{Sham Kakade}, \bibinfo{person}{Prateek Jain}, {and} \bibinfo{person}{Ali Farhadi}.} \bibinfo{year}{2022}\natexlab{}.
\newblock \showarticletitle{Matryoshka Representation Learning}.
\newblock \bibinfo{journal}{\emph{Advances in Neural Information Processing Systems}}  \bibinfo{volume}{35} (\bibinfo{year}{2022}), \bibinfo{pages}{30233--30249}.
\newblock


\bibitem[Lassance and Clinchant(2022)]%
        {formal2022efficientsplade}
\bibfield{author}{\bibinfo{person}{Carlos Lassance} {and} \bibinfo{person}{St\'{e}phane Clinchant}.} \bibinfo{year}{2022}\natexlab{}.
\newblock \showarticletitle{An Efficiency Study for SPLADE Models}. In \bibinfo{booktitle}{\emph{Proceedings of the 45th International ACM SIGIR Conference on Research and Development in Information Retrieval}} (Madrid, Spain) \emph{(\bibinfo{series}{SIGIR '22})}. \bibinfo{publisher}{Association for Computing Machinery}, \bibinfo{address}{New York, NY, USA}, \bibinfo{pages}{2220–2226}.
\newblock
\showISBNx{9781450387323}
\href{https://doi.org/10.1145/3477495.3531833}{doi:\nolinkurl{10.1145/3477495.3531833}}


\bibitem[Li et~al\mbox{.}(2024)]%
        {li20242dmatryoshkasentenceembeddings}
\bibfield{author}{\bibinfo{person}{Xianming Li}, \bibinfo{person}{Zongxi Li}, \bibinfo{person}{Jing Li}, \bibinfo{person}{Haoran Xie}, {and} \bibinfo{person}{Qing Li}.} \bibinfo{year}{2024}\natexlab{}.
\newblock \bibinfo{title}{2D Matryoshka Sentence Embeddings}.
\newblock
\showeprint[arxiv]{2402.14776}~[cs.CL]
\urldef\tempurl%
\url{https://arxiv.org/abs/2402.14776}
\showURL{%
\tempurl}


\bibitem[Lin et~al\mbox{.}(2022)]%
        {lin2022pretrained}
\bibfield{author}{\bibinfo{person}{Jimmy Lin}, \bibinfo{person}{Rodrigo Nogueira}, {and} \bibinfo{person}{Andrew Yates}.} \bibinfo{year}{2022}\natexlab{}.
\newblock \bibinfo{booktitle}{\emph{Pretrained Transformers for Text Ranking: {BERT} and Beyond}}.
\newblock \bibinfo{publisher}{Springer Nature}.
\newblock


\bibitem[Lin et~al\mbox{.}(2024)]%
        {lin2024awq}
\bibfield{author}{\bibinfo{person}{Ji Lin}, \bibinfo{person}{Jiaming Tang}, \bibinfo{person}{Haotian Tang}, \bibinfo{person}{Shang Yang}, \bibinfo{person}{Wei-Ming Chen}, \bibinfo{person}{Wei-Chen Wang}, \bibinfo{person}{Guangxuan Xiao}, \bibinfo{person}{Xingyu Dang}, \bibinfo{person}{Chuang Gan}, {and} \bibinfo{person}{Song Han}.} \bibinfo{year}{2024}\natexlab{}.
\newblock \showarticletitle{AWQ: Activation-aware Weight Quantization for On-Device LLM Compression and Acceleration}. In \bibinfo{booktitle}{\emph{Proceedings of Machine Learning and Systems}}, \bibfield{editor}{\bibinfo{person}{P.~Gibbons}, \bibinfo{person}{G.~Pekhimenko}, {and} \bibinfo{person}{C.~De Sa}} (Eds.), Vol.~\bibinfo{volume}{6}. \bibinfo{pages}{87--100}.
\newblock
\urldef\tempurl%
\url{https://proceedings.mlsys.org/paper_files/paper/2024/file/42a452cbafa9dd64e9ba4aa95cc1ef21-Paper-Conference.pdf}
\showURL{%
\tempurl}


\bibitem[Liu et~al\mbox{.}(2020)]%
        {liu-etal-2020-fastbert}
\bibfield{author}{\bibinfo{person}{Weijie Liu}, \bibinfo{person}{Peng Zhou}, \bibinfo{person}{Zhiruo Wang}, \bibinfo{person}{Zhe Zhao}, \bibinfo{person}{Haotang Deng}, {and} \bibinfo{person}{Qi Ju}.} \bibinfo{year}{2020}\natexlab{}.
\newblock \showarticletitle{{F}ast{BERT}: A Self-Distilling {BERT} with Adaptive Inference Time}. In \bibinfo{booktitle}{\emph{Proceedings of the 58th Annual Meeting of the Association for Computational Linguistics}}, \bibfield{editor}{\bibinfo{person}{Dan Jurafsky}, \bibinfo{person}{Joyce Chai}, \bibinfo{person}{Natalie Schluter}, {and} \bibinfo{person}{Joel Tetreault}} (Eds.). \bibinfo{publisher}{Association for Computational Linguistics}, \bibinfo{address}{Online}, \bibinfo{pages}{6035--6044}.
\newblock
\href{https://doi.org/10.18653/v1/2020.acl-main.537}{doi:\nolinkurl{10.18653/v1/2020.acl-main.537}}


\bibitem[Ma et~al\mbox{.}(2025)]%
        {ma2025tevatron20}
\bibfield{author}{\bibinfo{person}{Xueguang Ma}, \bibinfo{person}{Luyu Gao}, \bibinfo{person}{Shengyao Zhuang}, \bibinfo{person}{Jiaqi~Samantha Zhan}, \bibinfo{person}{Jamie Callan}, {and} \bibinfo{person}{Jimmy Lin}.} \bibinfo{year}{2025}\natexlab{}.
\newblock \showarticletitle{Tevatron 2.0: Unified Document Retrieval Toolkit across Scale, Language, and Modality}. In \bibinfo{booktitle}{\emph{Proceedings of the 48th International ACM SIGIR Conference on Research and Development in Information Retrieval}} (Padua, Italy) \emph{(\bibinfo{series}{SIGIR '25})}. \bibinfo{publisher}{Association for Computing Machinery}, \bibinfo{address}{New York, NY, USA}, \bibinfo{pages}{4061–4065}.
\newblock
\showISBNx{9798400715921}
\href{https://doi.org/10.1145/3726302.3730135}{doi:\nolinkurl{10.1145/3726302.3730135}}


\bibitem[Oord et~al\mbox{.}(2018)]%
        {oord2018representation}
\bibfield{author}{\bibinfo{person}{Aaron van~den Oord}, \bibinfo{person}{Yazhe Li}, {and} \bibinfo{person}{Oriol Vinyals}.} \bibinfo{year}{2018}\natexlab{}.
\newblock \bibinfo{title}{Representation Learning with Contrastive Predictive Coding}.
\newblock
\showeprint[arxiv]{1807.03748}~[cs.LG]
\urldef\tempurl%
\url{https://arxiv.org/abs/1807.03748}
\showURL{%
\tempurl}


\bibitem[Qu et~al\mbox{.}(2021)]%
        {qu-etal-2021-rocketqa}
\bibfield{author}{\bibinfo{person}{Yingqi Qu}, \bibinfo{person}{Yuchen Ding}, \bibinfo{person}{Jing Liu}, \bibinfo{person}{Kai Liu}, \bibinfo{person}{Ruiyang Ren}, \bibinfo{person}{Wayne~Xin Zhao}, \bibinfo{person}{Daxiang Dong}, \bibinfo{person}{Hua Wu}, {and} \bibinfo{person}{Haifeng Wang}.} \bibinfo{year}{2021}\natexlab{}.
\newblock \showarticletitle{{R}ocket{QA}: An Optimized Training Approach to Dense Passage Retrieval for Open-Domain Question Answering}. In \bibinfo{booktitle}{\emph{Proceedings of the 2021 Conference of the North American Chapter of the Association for Computational Linguistics: Human Language Technologies}}, \bibfield{editor}{\bibinfo{person}{Kristina Toutanova}, \bibinfo{person}{Anna Rumshisky}, \bibinfo{person}{Luke Zettlemoyer}, \bibinfo{person}{Dilek Hakkani-Tur}, \bibinfo{person}{Iz~Beltagy}, \bibinfo{person}{Steven Bethard}, \bibinfo{person}{Ryan Cotterell}, \bibinfo{person}{Tanmoy Chakraborty}, {and} \bibinfo{person}{Yichao Zhou}} (Eds.). \bibinfo{publisher}{Association for Computational Linguistics}, \bibinfo{address}{Online}, \bibinfo{pages}{5835--5847}.
\newblock
\href{https://doi.org/10.18653/v1/2021.naacl-main.466}{doi:\nolinkurl{10.18653/v1/2021.naacl-main.466}}


\bibitem[Sanh et~al\mbox{.}(2019)]%
        {sanh2019distilbert}
\bibfield{author}{\bibinfo{person}{Victor Sanh}, \bibinfo{person}{Lysandre Debut}, \bibinfo{person}{Julien Chaumond}, {and} \bibinfo{person}{Thomas Wolf}.} \bibinfo{year}{2019}\natexlab{}.
\newblock \bibinfo{title}{{DistilBERT}, a Distilled Version of {BERT}: Smaller, Faster, Cheaper and Lighter}.
\newblock
\showeprint[arxiv]{1910.01108}~[cs.CL]
\urldef\tempurl%
\url{https://arxiv.org/abs/1910.01108}
\showURL{%
\tempurl}


\bibitem[Thakur et~al\mbox{.}(2021)]%
        {thakur2021beir}
\bibfield{author}{\bibinfo{person}{Nandan Thakur}, \bibinfo{person}{Nils Reimers}, \bibinfo{person}{Andreas R{\"u}ckl{\'e}}, \bibinfo{person}{Abhishek Srivastava}, {and} \bibinfo{person}{Iryna Gurevych}.} \bibinfo{year}{2021}\natexlab{}.
\newblock \showarticletitle{{BEIR}: A Heterogeneous Benchmark for Zero-shot Evaluation of Information Retrieval Models}. In \bibinfo{booktitle}{\emph{Thirty-fifth Conference on Neural Information Processing Systems Datasets and Benchmarks Track (Round 2)}}.
\newblock
\urldef\tempurl%
\url{https://openreview.net/forum?id=wCu6T5xFjeJ}
\showURL{%
\tempurl}


\bibitem[Thakur et~al\mbox{.}(2025)]%
        {thakur-etal-2025-hard}
\bibfield{author}{\bibinfo{person}{Nandan Thakur}, \bibinfo{person}{Crystina Zhang}, \bibinfo{person}{Xueguang Ma}, {and} \bibinfo{person}{Jimmy Lin}.} \bibinfo{year}{2025}\natexlab{}.
\newblock \showarticletitle{Hard Negatives, Hard Lessons: Revisiting Training Data Quality for Robust Information Retrieval with {LLM}s}. In \bibinfo{booktitle}{\emph{Findings of the Association for Computational Linguistics: EMNLP 2025}}, \bibfield{editor}{\bibinfo{person}{Christos Christodoulopoulos}, \bibinfo{person}{Tanmoy Chakraborty}, \bibinfo{person}{Carolyn Rose}, {and} \bibinfo{person}{Violet Peng}} (Eds.). \bibinfo{publisher}{Association for Computational Linguistics}, \bibinfo{address}{Suzhou, China}, \bibinfo{pages}{9064--9083}.
\newblock
\showISBNx{979-8-89176-335-7}
\href{https://doi.org/10.18653/v1/2025.findings-emnlp.481}{doi:\nolinkurl{10.18653/v1/2025.findings-emnlp.481}}


\bibitem[Warner et~al\mbox{.}(2024)]%
        {warner2024modernbert}
\bibfield{author}{\bibinfo{person}{Benjamin Warner}, \bibinfo{person}{Antoine Chaffin}, \bibinfo{person}{Benjamin Clavi{\'e}}, \bibinfo{person}{Orion Weller}, \bibinfo{person}{Oskar Hallstr{\"o}m}, \bibinfo{person}{Said Taghadouini}, \bibinfo{person}{Alexis Gallagher}, \bibinfo{person}{Raja Biswas}, \bibinfo{person}{Faisal Ladhak}, \bibinfo{person}{Tom Aarsen}, \bibinfo{person}{Nathan Cooper}, \bibinfo{person}{Griffin Adams}, \bibinfo{person}{Jeremy Howard}, {and} \bibinfo{person}{Iacopo Poli}.} \bibinfo{year}{2024}\natexlab{}.
\newblock \bibinfo{title}{Smarter, Better, Faster, Longer: A Modern Bidirectional Encoder for Fast, Memory Efficient, and Long Context Finetuning and Inference}.
\newblock
\showeprint[arxiv]{2412.13663}~[cs.CL]
\urldef\tempurl%
\url{https://arxiv.org/abs/2412.13663}
\showURL{%
\tempurl}


\bibitem[Xia et~al\mbox{.}(2008)]%
        {xia2008listwiseapproach}
\bibfield{author}{\bibinfo{person}{Fen Xia}, \bibinfo{person}{Tie-Yan Liu}, \bibinfo{person}{Jue Wang}, \bibinfo{person}{Wensheng Zhang}, {and} \bibinfo{person}{Hang Li}.} \bibinfo{year}{2008}\natexlab{}.
\newblock \showarticletitle{Listwise Approach to Learning to Rank: Theory and Algorithm}. In \bibinfo{booktitle}{\emph{Proceedings of the 25th International Conference on Machine Learning}} (Helsinki, Finland) \emph{(\bibinfo{series}{ICML '08})}. \bibinfo{publisher}{Association for Computing Machinery}, \bibinfo{address}{New York, NY, USA}, \bibinfo{pages}{1192–1199}.
\newblock
\showISBNx{9781605582054}
\href{https://doi.org/10.1145/1390156.1390306}{doi:\nolinkurl{10.1145/1390156.1390306}}


\bibitem[Xia et~al\mbox{.}(2024)]%
        {xia2024shearedllama}
\bibfield{author}{\bibinfo{person}{Mengzhou Xia}, \bibinfo{person}{Tianyu Gao}, \bibinfo{person}{Zhiyuan Zeng}, {and} \bibinfo{person}{Danqi Chen}.} \bibinfo{year}{2024}\natexlab{}.
\newblock \showarticletitle{Sheared {LLaMA}: Accelerating Language Model Pre-training via Structured Pruning}. In \bibinfo{booktitle}{\emph{International Conference on Learning Representations (ICLR)}}.
\newblock


\bibitem[Xin et~al\mbox{.}(2020)]%
        {xin-etal-2020-deebert}
\bibfield{author}{\bibinfo{person}{Ji Xin}, \bibinfo{person}{Raphael Tang}, \bibinfo{person}{Jaejun Lee}, \bibinfo{person}{Yaoliang Yu}, {and} \bibinfo{person}{Jimmy Lin}.} \bibinfo{year}{2020}\natexlab{}.
\newblock \showarticletitle{{D}ee{BERT}: Dynamic Early Exiting for Accelerating {BERT} Inference}. In \bibinfo{booktitle}{\emph{Proceedings of the 58th Annual Meeting of the Association for Computational Linguistics}}, \bibfield{editor}{\bibinfo{person}{Dan Jurafsky}, \bibinfo{person}{Joyce Chai}, \bibinfo{person}{Natalie Schluter}, {and} \bibinfo{person}{Joel Tetreault}} (Eds.). \bibinfo{publisher}{Association for Computational Linguistics}, \bibinfo{address}{Online}, \bibinfo{pages}{2246--2251}.
\newblock
\href{https://doi.org/10.18653/v1/2020.acl-main.204}{doi:\nolinkurl{10.18653/v1/2020.acl-main.204}}


\bibitem[Xiong et~al\mbox{.}(2021)]%
        {xiong2021ance}
\bibfield{author}{\bibinfo{person}{Lee Xiong}, \bibinfo{person}{Chenyan Xiong}, \bibinfo{person}{Ye Li}, \bibinfo{person}{Kwok-Fung Tang}, \bibinfo{person}{Jialin Liu}, \bibinfo{person}{Paul~N. Bennett}, \bibinfo{person}{Junaid Ahmed}, {and} \bibinfo{person}{Arnold Overwijk}.} \bibinfo{year}{2021}\natexlab{}.
\newblock \showarticletitle{Approximate Nearest Neighbor Negative Contrastive Learning for Dense Text Retrieval}. In \bibinfo{booktitle}{\emph{International Conference on Learning Representations}}.
\newblock
\urldef\tempurl%
\url{https://openreview.net/forum?id=zeFrfgyZln}
\showURL{%
\tempurl}


\bibitem[Xu(2024)]%
        {xu2024rankmambabenchmarkingmambasdocument}
\bibfield{author}{\bibinfo{person}{Zhichao Xu}.} \bibinfo{year}{2024}\natexlab{}.
\newblock \bibinfo{title}{RankMamba: Benchmarking Mamba's Document Ranking Performance in the Era of Transformers}.
\newblock
\showeprint[arxiv]{2403.18276}~[cs.IR]
\urldef\tempurl%
\url{https://arxiv.org/abs/2403.18276}
\showURL{%
\tempurl}


\bibitem[Xu et~al\mbox{.}(2025a)]%
        {xu2025cspladelearnedsparseretrieval}
\bibfield{author}{\bibinfo{person}{Zhichao Xu}, \bibinfo{person}{Aosong Feng}, \bibinfo{person}{Yijun Tian}, \bibinfo{person}{Haibo Ding}, {and} \bibinfo{person}{Lin~Lee Cheong}.} \bibinfo{year}{2025}\natexlab{a}.
\newblock \showarticletitle{{CSPLADE}: Learned Sparse Retrieval with Causal Language Models}. In \bibinfo{booktitle}{\emph{Proceedings of the 14th International Joint Conference on Natural Language Processing and the 4th Conference of the Asia-Pacific Chapter of the Association for Computational Linguistics}}, \bibfield{editor}{\bibinfo{person}{Kentaro Inui}, \bibinfo{person}{Sakriani Sakti}, \bibinfo{person}{Haofen Wang}, \bibinfo{person}{Derek~F. Wong}, \bibinfo{person}{Pushpak Bhattacharyya}, \bibinfo{person}{Biplab Banerjee}, \bibinfo{person}{Asif Ekbal}, \bibinfo{person}{Tanmoy Chakraborty}, {and} \bibinfo{person}{Dhirendra~Pratap Singh}} (Eds.). \bibinfo{publisher}{The Asian Federation of Natural Language Processing and The Association for Computational Linguistics}, \bibinfo{address}{Mumbai, India}, \bibinfo{pages}{99--114}.
\newblock
\showISBNx{979-8-89176-298-5}
\href{https://doi.org/10.18653/v1/2025.ijcnlp-long.7}{doi:\nolinkurl{10.18653/v1/2025.ijcnlp-long.7}}


\bibitem[Xu et~al\mbox{.}(2024)]%
        {xu-etal-2024-beyond-perplexity}
\bibfield{author}{\bibinfo{person}{Zhichao Xu}, \bibinfo{person}{Ashim Gupta}, \bibinfo{person}{Tao Li}, \bibinfo{person}{Oliver Bentham}, {and} \bibinfo{person}{Vivek Srikumar}.} \bibinfo{year}{2024}\natexlab{}.
\newblock \showarticletitle{Beyond Perplexity: Multi-dimensional Safety Evaluation of {LLM} Compression}. In \bibinfo{booktitle}{\emph{Findings of the Association for Computational Linguistics: EMNLP 2024}}, \bibfield{editor}{\bibinfo{person}{Yaser Al-Onaizan}, \bibinfo{person}{Mohit Bansal}, {and} \bibinfo{person}{Yun-Nung Chen}} (Eds.). \bibinfo{publisher}{Association for Computational Linguistics}, \bibinfo{address}{Miami, Florida, USA}, \bibinfo{pages}{15359--15396}.
\newblock
\href{https://doi.org/10.18653/v1/2024.findings-emnlp.901}{doi:\nolinkurl{10.18653/v1/2024.findings-emnlp.901}}


\bibitem[Xu et~al\mbox{.}(2025b)]%
        {xu2025distillationversuscontrastivelearning}
\bibfield{author}{\bibinfo{person}{Zhichao Xu}, \bibinfo{person}{Zhiqi Huang}, \bibinfo{person}{Shengyao Zhuang}, {and} \bibinfo{person}{Vivek Srikumar}.} \bibinfo{year}{2025}\natexlab{b}.
\newblock \showarticletitle{Distillation versus Contrastive Learning: How to Train Your Rerankers}. In \bibinfo{booktitle}{\emph{Proceedings of the 14th International Joint Conference on Natural Language Processing and the 4th Conference of the Asia-Pacific Chapter of the Association for Computational Linguistics}}, \bibfield{editor}{\bibinfo{person}{Kentaro Inui}, \bibinfo{person}{Sakriani Sakti}, \bibinfo{person}{Haofen Wang}, \bibinfo{person}{Derek~F. Wong}, \bibinfo{person}{Pushpak Bhattacharyya}, \bibinfo{person}{Biplab Banerjee}, \bibinfo{person}{Asif Ekbal}, \bibinfo{person}{Tanmoy Chakraborty}, {and} \bibinfo{person}{Dhirendra~Pratap Singh}} (Eds.). \bibinfo{publisher}{The Asian Federation of Natural Language Processing and The Association for Computational Linguistics}, \bibinfo{address}{Mumbai, India}, \bibinfo{pages}{564--578}.
\newblock
\showISBNx{979-8-89176-303-6}
\href{https://doi.org/10.18653/v1/2025.findings-ijcnlp.33}{doi:\nolinkurl{10.18653/v1/2025.findings-ijcnlp.33}}


\bibitem[Xu et~al\mbox{.}(2026a)]%
        {xu2026tevatronmeetsmegatronexpertparallel}
\bibfield{author}{\bibinfo{person}{Zhichao Xu}, \bibinfo{person}{Xueguang Ma}, \bibinfo{person}{Shengyao Zhuang}, \bibinfo{person}{Luyu Gao}, \bibinfo{person}{Wenqian Ye}, \bibinfo{person}{Yu Wang}, \bibinfo{person}{Jamie Callan}, {and} \bibinfo{person}{Jimmy Lin}.} \bibinfo{year}{2026}\natexlab{a}.
\newblock \bibinfo{title}{Tevatron Meets Megatron: Expert-Parallel LLM Reranker Training on an Academic Budget}.
\newblock
\showeprint[arxiv]{2608.00916}~[cs.IR]
\urldef\tempurl%
\url{https://arxiv.org/abs/2608.00916}
\showURL{%
\tempurl}


\bibitem[Xu et~al\mbox{.}(2026b)]%
        {xu2025surveymodelarchitecturesinformation}
\bibfield{author}{\bibinfo{person}{Zhichao Xu}, \bibinfo{person}{Fengran Mo}, \bibinfo{person}{Zhiqi Huang}, \bibinfo{person}{Crystina Zhang}, \bibinfo{person}{Puxuan Yu}, \bibinfo{person}{Bei~Wang Phillips}, \bibinfo{person}{Jimmy Lin}, {and} \bibinfo{person}{Vivek Srikumar}.} \bibinfo{year}{2026}\natexlab{b}.
\newblock \showarticletitle{A Survey of Model Architectures in Information Retrieval}.
\newblock \bibinfo{journal}{\emph{Transactions on Machine Learning Research}} (\bibinfo{year}{2026}).
\newblock
\shownote{Survey Certification}.
\newblock
\showISSN{2835-8856}
\urldef\tempurl%
\url{https://openreview.net/forum?id=xAIbTbHRrX}
\showURL{%
\tempurl}


\bibitem[Xu et~al\mbox{.}(2026c)]%
        {xu2026reconreasoningcondensationefficient}
\bibfield{author}{\bibinfo{person}{Zhichao Xu}, \bibinfo{person}{Minheng Wang}, \bibinfo{person}{Yawei Wang}, \bibinfo{person}{Wenqian Ye}, \bibinfo{person}{Yuntao Du}, \bibinfo{person}{Yunpu Ma}, {and} \bibinfo{person}{Yijun Tian}.} \bibinfo{year}{2026}\natexlab{c}.
\newblock \bibinfo{title}{RECON: Reasoning with Condensation for Efficient Retrieval-Augmented Generation}.
\newblock
\showeprint[arxiv]{2510.10448}~[cs.CL]
\urldef\tempurl%
\url{https://arxiv.org/abs/2510.10448}
\showURL{%
\tempurl}


\bibitem[Xu et~al\mbox{.}(2026d)]%
        {xu2026beyond}
\bibfield{author}{\bibinfo{person}{Zhichao Xu}, \bibinfo{person}{Zongyu Wu}, \bibinfo{person}{Yun Zhou}, \bibinfo{person}{Aosong Feng}, \bibinfo{person}{Kang Zhou}, \bibinfo{person}{Sangmin Woo}, \bibinfo{person}{Kiran Ramnath}, \bibinfo{person}{Yijun Tian}, \bibinfo{person}{Xuan Qi}, \bibinfo{person}{Weikang Qiu}, \bibinfo{person}{Lin~Lee Cheong}, {and} \bibinfo{person}{Haibo Ding}.} \bibinfo{year}{2026}\natexlab{d}.
\newblock \showarticletitle{Beyond Correctness: Rewarding Faithful Reasoning in Retrieval-Augmented Generation}.
\newblock \bibinfo{journal}{\emph{Transactions on Machine Learning Research}} (\bibinfo{year}{2026}).
\newblock
\showISSN{2835-8856}
\urldef\tempurl%
\url{https://openreview.net/forum?id=mZ0gGlXelF}
\showURL{%
\tempurl}


\bibitem[Xu et~al\mbox{.}(2025c)]%
        {xu-etal-2025-state}
\bibfield{author}{\bibinfo{person}{Zhichao Xu}, \bibinfo{person}{Jinghua Yan}, \bibinfo{person}{Ashim Gupta}, {and} \bibinfo{person}{Vivek Srikumar}.} \bibinfo{year}{2025}\natexlab{c}.
\newblock \showarticletitle{State Space Models are Strong Text Rerankers}. In \bibinfo{booktitle}{\emph{Proceedings of the 10th Workshop on Representation Learning for NLP (RepL4NLP-2025)}}, \bibfield{editor}{\bibinfo{person}{Vaibhav Adlakha}, \bibinfo{person}{Alexandra Chronopoulou}, \bibinfo{person}{Xiang~Lorraine Li}, \bibinfo{person}{Bodhisattwa~Prasad Majumder}, \bibinfo{person}{Freda Shi}, {and} \bibinfo{person}{Giorgos Vernikos}} (Eds.). \bibinfo{publisher}{Association for Computational Linguistics}, \bibinfo{address}{Albuquerque, NM}, \bibinfo{pages}{152--169}.
\newblock
\showISBNx{979-8-89176-245-9}
\href{https://doi.org/10.18653/v1/2025.repl4nlp-1.12}{doi:\nolinkurl{10.18653/v1/2025.repl4nlp-1.12}}


\bibitem[Xu et~al\mbox{.}(2026e)]%
        {xu2026rethinking}
\bibfield{author}{\bibinfo{person}{Zhichao Xu}, \bibinfo{person}{Shengyao Zhuang}, \bibinfo{person}{Xueguang Ma}, \bibinfo{person}{Bingsen Chen}, \bibinfo{person}{Yijun Tian}, \bibinfo{person}{Fengran Mo}, \bibinfo{person}{Tao Li}, \bibinfo{person}{Jie Cao}, {and} \bibinfo{person}{Vivek Srikumar}.} \bibinfo{year}{2026}\natexlab{e}.
\newblock \showarticletitle{Rethinking On-policy Optimization for Query Augmentation}.
\newblock \bibinfo{journal}{\emph{Transactions on Machine Learning Research}} (\bibinfo{year}{2026}).
\newblock
\showISSN{2835-8856}
\urldef\tempurl%
\url{https://openreview.net/forum?id=mmqbjhz5Br}
\showURL{%
\tempurl}


\bibitem[Xu et~al\mbox{.}(2026f)]%
        {xu2026laconic}
\bibfield{author}{\bibinfo{person}{Zhichao Xu}, \bibinfo{person}{Shengyao Zhuang}, \bibinfo{person}{Crystina Zhang}, \bibinfo{person}{Xueguang Ma}, \bibinfo{person}{Yijun Tian}, \bibinfo{person}{Maitrey Mehta}, \bibinfo{person}{Jimmy Lin}, {and} \bibinfo{person}{Vivek Srikumar}.} \bibinfo{year}{2026}\natexlab{f}.
\newblock \showarticletitle{LACONIC: Dense-Level Effectiveness for Scalable Sparse Retrieval via a Two-Phase Training Curriculum}. In \bibinfo{booktitle}{\emph{Proceedings of the 49th International ACM SIGIR Conference on Research and Development in Information Retrieval}} (Australia) \emph{(\bibinfo{series}{SIGIR '26})}. \bibinfo{publisher}{Association for Computing Machinery}, \bibinfo{address}{New York, NY, USA}, \bibinfo{pages}{4298–4304}.
\newblock
\showISBNx{9798400725999}
\href{https://doi.org/10.1145/3805712.3809869}{doi:\nolinkurl{10.1145/3805712.3809869}}


\bibitem[Yang et~al\mbox{.}(2025)]%
        {yang2025qwen3technicalreport}
\bibfield{author}{\bibinfo{person}{An Yang}, \bibinfo{person}{Anfeng Li}, \bibinfo{person}{Baosong Yang}, \bibinfo{person}{Beichen Zhang}, \bibinfo{person}{Binyuan Hui}, \bibinfo{person}{Bo Zheng}, \bibinfo{person}{Bowen Yu}, \bibinfo{person}{Chang Gao}, \bibinfo{person}{Chengen Huang}, \bibinfo{person}{Chenxu Lv}, \bibinfo{person}{Chujie Zheng}, \bibinfo{person}{Dayiheng Liu}, \bibinfo{person}{Fan Zhou}, \bibinfo{person}{Fei Huang}, \bibinfo{person}{Feng Hu}, \bibinfo{person}{Hao Ge}, \bibinfo{person}{Haoran Wei}, \bibinfo{person}{Huan Lin}, \bibinfo{person}{Jialong Tang}, \bibinfo{person}{Jian Yang}, \bibinfo{person}{Jianhong Tu}, \bibinfo{person}{Jianwei Zhang}, \bibinfo{person}{Jianxin Yang}, \bibinfo{person}{Jiaxi Yang}, \bibinfo{person}{Jing Zhou}, \bibinfo{person}{Jingren Zhou}, \bibinfo{person}{Junyang Lin}, \bibinfo{person}{Kai Dang}, \bibinfo{person}{Keqin Bao}, \bibinfo{person}{Kexin Yang}, \bibinfo{person}{Le Yu}, \bibinfo{person}{Lianghao Deng}, \bibinfo{person}{Mei Li}, \bibinfo{person}{Mingfeng
  Xue}, \bibinfo{person}{Mingze Li}, \bibinfo{person}{Pei Zhang}, \bibinfo{person}{Peng Wang}, \bibinfo{person}{Qin Zhu}, \bibinfo{person}{Rui Men}, \bibinfo{person}{Ruize Gao}, \bibinfo{person}{Shixuan Liu}, \bibinfo{person}{Shuang Luo}, \bibinfo{person}{Tianhao Li}, \bibinfo{person}{Tianyi Tang}, \bibinfo{person}{Wenbiao Yin}, \bibinfo{person}{Xingzhang Ren}, \bibinfo{person}{Xinyu Wang}, \bibinfo{person}{Xinyu Zhang}, \bibinfo{person}{Xuancheng Ren}, \bibinfo{person}{Yang Fan}, \bibinfo{person}{Yang Su}, \bibinfo{person}{Yichang Zhang}, \bibinfo{person}{Yinger Zhang}, \bibinfo{person}{Yu Wan}, \bibinfo{person}{Yuqiong Liu}, \bibinfo{person}{Zekun Wang}, \bibinfo{person}{Zeyu Cui}, \bibinfo{person}{Zhenru Zhang}, \bibinfo{person}{Zhipeng Zhou}, {and} \bibinfo{person}{Zihan Qiu}.} \bibinfo{year}{2025}\natexlab{}.
\newblock \bibinfo{title}{Qwen3 Technical Report}.
\newblock
\showeprint[arxiv]{2505.09388}~[cs.CL]
\urldef\tempurl%
\url{https://arxiv.org/abs/2505.09388}
\showURL{%
\tempurl}


\bibitem[Yu et~al\mbox{.}(2019)]%
        {yu2019slimmable}
\bibfield{author}{\bibinfo{person}{Jiahui Yu}, \bibinfo{person}{Linjie Yang}, \bibinfo{person}{Ning Xu}, \bibinfo{person}{Jianchao Yang}, {and} \bibinfo{person}{Thomas Huang}.} \bibinfo{year}{2019}\natexlab{}.
\newblock \showarticletitle{Slimmable Neural Networks}. In \bibinfo{booktitle}{\emph{International Conference on Learning Representations (ICLR)}}.
\newblock


\bibitem[Zhang et~al\mbox{.}(2025)]%
        {zhang2025jaspertokencompression600mtechnicalreport}
\bibfield{author}{\bibinfo{person}{Dun Zhang}, \bibinfo{person}{Ziyang Zeng}, \bibinfo{person}{Yudong Zhou}, {and} \bibinfo{person}{Shuyang Lu}.} \bibinfo{year}{2025}\natexlab{}.
\newblock \bibinfo{title}{Jasper-Token-Compression-600M Technical Report}.
\newblock
\showeprint[arxiv]{2511.14405}~[cs.IR]
\urldef\tempurl%
\url{https://arxiv.org/abs/2511.14405}
\showURL{%
\tempurl}


\bibitem[Zhuang et~al\mbox{.}(2026a)]%
        {zhuang2026starbucks}
\bibfield{author}{\bibinfo{person}{Shengyao Zhuang}, \bibinfo{person}{Shuai Wang}, \bibinfo{person}{Fabio Zheng}, \bibinfo{person}{Bevan Koopman}, {and} \bibinfo{person}{Guido Zuccon}.} \bibinfo{year}{2026}\natexlab{a}.
\newblock \showarticletitle{Starbucks: Improved Training for 2D Matryoshka Embeddings}. In \bibinfo{booktitle}{\emph{Advances in Information Retrieval: 48th European Conference on Information Retrieval, ECIR 2026, Delft, The Netherlands, March 29 – April 2, 2026, Proceedings, Part I}} (Delft, The Netherlands). \bibinfo{publisher}{Springer-Verlag}, \bibinfo{address}{Berlin, Heidelberg}, \bibinfo{pages}{67–82}.
\newblock
\showISBNx{978-3-032-21288-7}
\href{https://doi.org/10.1007/978-3-032-21289-4_5}{doi:\nolinkurl{10.1007/978-3-032-21289-4_5}}


\bibitem[Zhuang et~al\mbox{.}(2026b)]%
        {Zhuang2026LayerwiseTC}
\bibfield{author}{\bibinfo{person}{Shengyao Zhuang}, \bibinfo{person}{Zhichao Xu}, {and} \bibinfo{person}{Ivano Lauriola}.} \bibinfo{year}{2026}\natexlab{b}.
\newblock \showarticletitle{Layer-wise Token Compression for Efficient Document Reranking}. In \bibinfo{booktitle}{\emph{Proceedings of the 49th International ACM SIGIR Conference on Research and Development in Information Retrieval}} (Australia) \emph{(\bibinfo{series}{SIGIR '26})}. \bibinfo{publisher}{Association for Computing Machinery}, \bibinfo{address}{New York, NY, USA}, \bibinfo{pages}{4426–4432}.
\newblock
\showISBNx{9798400725999}
\href{https://doi.org/10.1145/3805712.3809871}{doi:\nolinkurl{10.1145/3805712.3809871}}


\end{thebibliography}
